\documentclass[journal]{IEEEtran}
\ifCLASSINFOpdf
\else
\fi
\usepackage{color}
\usepackage{graphicx} 
\usepackage{makecell}

\newcommand{\eg}{\textit{e}.\textit{g}.}

\newcommand{\fref}[1]{Fig.~\ref{#1}}

\begin{document}
%
% paper title
% Titles are generally capitalized except for words such as a, an, and, as,
% at, but, by, for, in, nor, of, on, or, the, to and up, which are usually
% not capitalized unless they are the first or last word of the title.
% Linebreaks \\ can be used within to get better formatting as desired.
% Do not put math or special symbols in the title.
\title{Edge Preserving and Multi-Scale Contextual Neural Network for Salient Object Detection}
%
%
% author names and IEEE memberships
% note positions of commas and nonbreaking spaces ( ~ ) LaTeX will not break
% a structure at a ~ so this keeps an author's name from being broken across
% two lines.
% use \thanks{} to gain access to the first footnote area
% a separate \thanks must be used for each paragraph as LaTeX2e's \thanks
% was not built to handle multiple paragraphs
%

\author{Xiang~Wang,
        Huimin~Ma,
        Xiaozhi~Chen,
        and~Shaodi~You% <-this % stops a space
\thanks{X. Wang, H. Ma and X. Chen are with Tsinghua National Laboratory for Information Science and Technology (TNList)
and Department of Electronic Engineering, Tsinghua University, Beijing 100084, China. E-mail: wangxiang14@mails.tsinghua.edu.cn, mhmpub@tsinghua.edu.cn, chenxz12@mails.tsinghua.edu.cn.}% <-this % stops a space
\thanks{S. You is with Data61, CSIRO and Australian National University, Australia. E-mail: Shaodi.You@data61.csiro.au.}
\thanks{The corresponding author is H. Ma.}}
%\thanks{Manuscript received April 19, 2005; revised August 26, 2015.}}

% note the % following the last \IEEEmembership and also \thanks - 
% these prevent an unwanted space from occurring between the last author name
% and the end of the author line. i.e., if you had this:
% 
% \author{....lastname \thanks{...} \thanks{...} }
%                     ^------------^------------^----Do not want these spaces!
%
% a space would be appended to the last name and could cause every name on that
% line to be shifted left slightly. This is one of those "LaTeX things". For
% instance, "\textbf{A} \textbf{B}" will typeset as "A B" not "AB". To get
% "AB" then you have to do: "\textbf{A}\textbf{B}"
% \thanks is no different in this regard, so shield the last } of each \thanks
% that ends a line with a % and do not let a space in before the next \thanks.
% Spaces after \IEEEmembership other than the last one are OK (and needed) as
% you are supposed to have spaces between the names. For what it is worth,
% this is a minor point as most people would not even notice if the said evil
% space somehow managed to creep in.

% The paper headers
\markboth{Journal of \LaTeX\ Class Files,~Vol.~14, No.~8, August~2015}%
{Shell \MakeLowercase{\textit{et al.}}: Bare Demo of IEEEtran.cls for IEEE Journals}
% The only time the second header will appear is for the odd numbered pages
% after the title page when using the twoside option.
% 
% *** Note that you probably will NOT want to include the author's ***
% *** name in the headers of peer review papers.                   ***
% You can use \ifCLASSOPTIONpeerreview for conditional compilation here if
% you desire.

% If you want to put a publisher's ID mark on the page you can do it like
% this:
%\IEEEpubid{0000--0000/00\$00.00~\copyright~2015 IEEE}
% Remember, if you use this you must call \IEEEpubidadjcol in the second
% column for its text to clear the IEEEpubid mark.

% use for special paper notices
%\IEEEspecialpapernotice{(Invited Paper)}

% make the title area
\maketitle

% As a general rule, do not put math, special symbols or citations
% in the abstract or keywords.
\begin{abstract}
In this paper, we propose a novel edge preserving and multi-scale contextual neural network for salient object detection. 
The proposed framework is aiming to address two limits of the existing CNN based methods. 
First, region-based CNN methods lack sufficient context to accurately locate salient object since they deal with each region independently.
Second, pixel-based CNN methods suffer from blurry boundaries due to the presence of convolutional and pooling layers. 
Motivated by these, we first propose an end-to-end edge-preserved neural network based on Fast R-CNN framework (named \textit{RegionNet}) to efficiently generate saliency map with sharp object boundaries. Later, to further improve it, multi-scale spatial context is attached to \textit{RegionNet} to consider the relationship between regions and the global scenes. Furthermore, our method can be generally applied to RGB-D saliency detection by depth refinement. The proposed framework achieves both clear detection boundary and multi-scale contextual robustness simultaneously for the first time, and thus achieves an optimized performance. Experiments on six RGB and two RGB-D benchmark datasets demonstrate that the proposed method achieves state-of-the-art performance.

\end{abstract}

% Note that keywords are not normally used for peerreview papers.

% For peer review papers, you can put extra information on the cover
% page as needed:
% \ifCLASSOPTIONpeerreview
% \begin{center} \bfseries EDICS Category: 3-BBND \end{center}
% \fi
%
% For peerreview papers, this IEEEtran command inserts a page break and
% creates the second title. It will be ignored for other modes.
\IEEEpeerreviewmaketitle

\section{Introduction}\label{sec:intro}
\IEEEPARstart{S}{alient} object detection, which aims to detect object that most attracts people's attention through out an image, has been widely exploited in recent years. It has also been widely utilized for many computer vision tasks, such as semantic segmentation~\cite{wei2015stc}, object tracking~\cite{mahadevan2009saliency, hong2015online} and image classification~\cite{lei2015saliency, 7270295}. 

Traditional saliency methods aim to generate a heat map which gives each pixel a relative value of its level of saliency \cite{itti1998model,zhang2008sun,murray2011saliency}. In recent years, the fashion moves to salient object detection which generates pixel-wise binary label for salient and non-salient objects \cite{zhai2006visual, cheng2011global, borji2014salient}. In comparing with the heat map, the binary label would further benefit segmentation based applications such as semantic segmentation~\cite{wei2015stc}, and thus attracts more attention.

To achieve a high accuracy for binary labeling, there are mainly two requirements:  first, multi-scale contextual reliability; and second, sharp boundary between salient and non-salient objects. The contextual reliability aims to  model the relationship between regions and global scenes to determine which object is salient. And the clear boundary aims to separate the salient object and background clearly and to highlight the whole object uniformly.

Unfortunately, none of the existing methods achieve both requirements simultaneously.
Traditional bottom-up methods mainly rely on priors or assumptions and hand-crafted features. For example, center-surround difference~\cite{itti1998model,liu2007learning}, uniqueness prior~\cite{shi2013pisa, jiang2013salient} and backgroundness prior~\cite{wei2012geodesic, zhu2014saliency}. These methods can not consider high-level semantic contextual relations and do not achieve a satisfying accuracy.

\begin{figure}[!t] %\footnotesize
\begin{center}
\includegraphics[width=1\linewidth,trim = 0mm 0mm 0mm 0mm, clip]{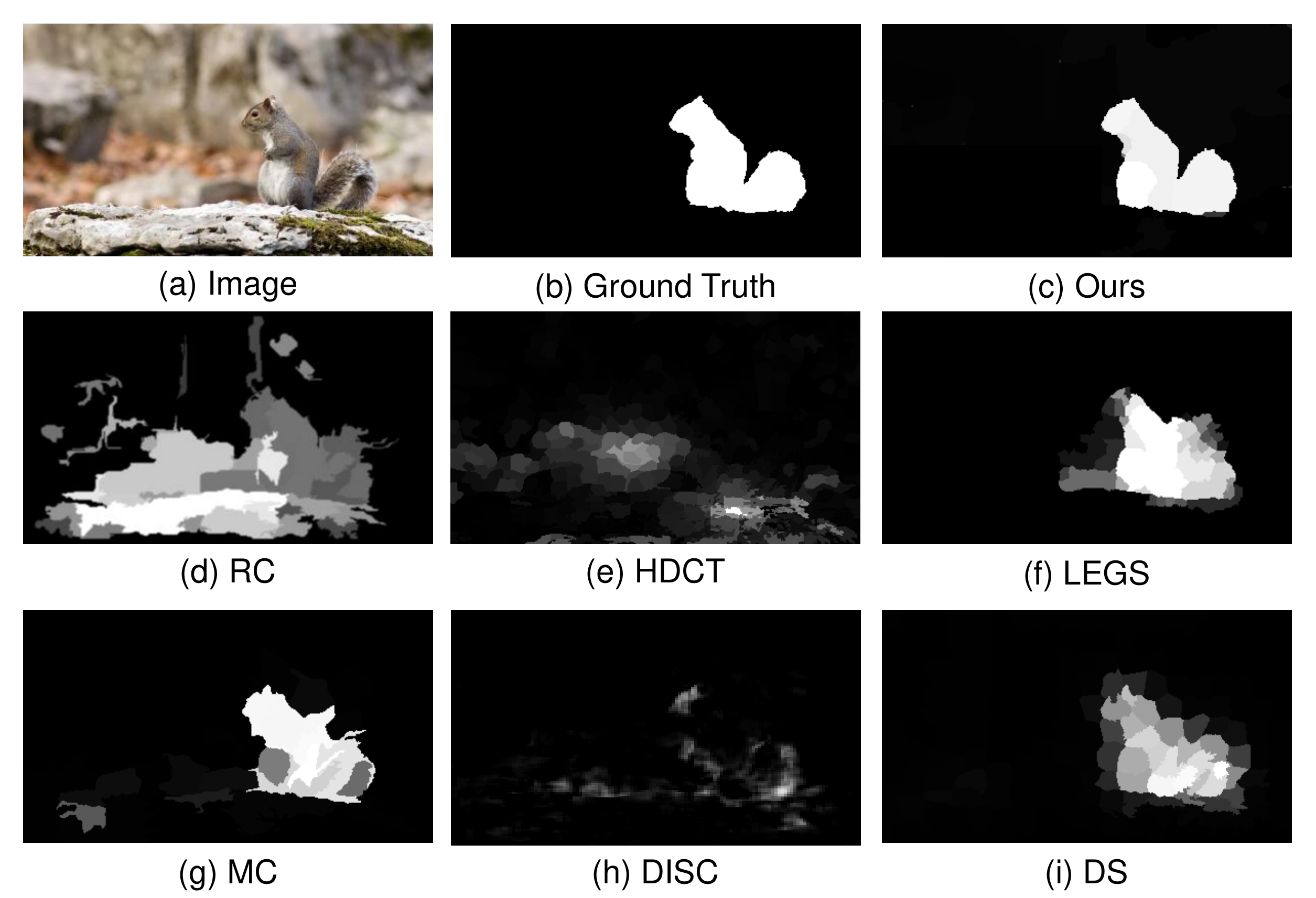}
\vspace{-5mm}
\caption{Saliency map of an image with low-contrast. Previous methods fail to distinguish the object from the confusing background. Our method detect salient object with fine boundaries by taking advantages of regions and multi-scale context. (a) image, (b) groundtruth, (c) our proposed \textit{RexNet}. (d, e) traditional methods: RC~\cite{cheng2011global} and HDCT~\cite{kim2014salient}, (f, g) region-based CNN methods: LEGS~\cite{wang2015deep} and MC~\cite{zhao2015saliency}, (h, i) pixel-based CNN methods: DISC~\cite{chen2015disc} and DS~\cite{li2016deepsaliency}.}

\label{fig:impressive}
\end{center}
\vspace{-5mm}
\end{figure}

\begin{figure*}[!t] %\footnotesize
\begin{center}
\includegraphics[width=1\linewidth,trim = 0mm 0mm 0mm 0mm, clip]{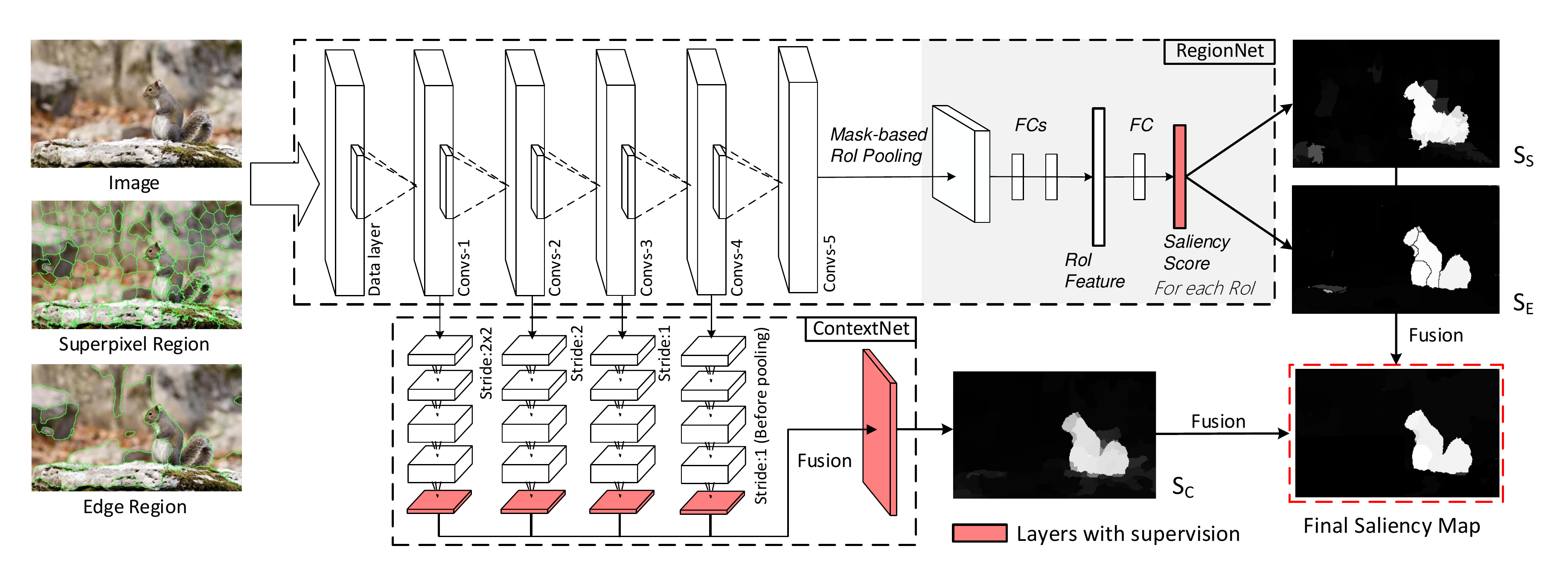}
\vspace{-10mm}
\caption{Architecture of the proposed \textit{RexNet}. The network is composed by two components: \textit{RegionNet} and \textit{ContextNet}. Image is first segmented into regions using superpixel and edges. \textit{RegionNet} predicts saliency score of regions and forms saliency maps $S_S$ and $S_E$. At the same time, \textit{ContextNet} extracts multi-scale spatial context and fuse them to get saliency map $S_C$. These three saliency maps are fused to get the final saliency map.}
\label{fig:framework}
\end{center}
\vspace{-5mm}
\end{figure*} 
Recently, the deep Convolutional Neural Network (CNN) has attracted wide attention for its superior performance. CNN based methods can be divided into region-based networks and pixel-based networks. Region-based methods aim to extract features of each region (or patch), and then predict its saliency score. However, existing region-based methods lack of representing context information to model the relationship between regions and global scenes. Because of this, it may have false detection results when the scene is complex or the object is composed by several different parts, which limits their performance (\fref{fig:impressive}).
On the other hand, existing pixel-based CNN methods lack the ability to produce clear boundary between salient and non-salient objects, due to the presence of convolutional and pooling layers, and they only achieve partial contextual reliability. This limits the performance of pixel-based methods (\fref{fig:impressive}).

In this paper, we propose a novel edge preserving and multi-scale contextual network for salient object detection. The proposed framework achieves both clear boundary and multi-scale contextual robustness simultaneously for the first time.
As illustrated in \fref{fig:framework}, the proposed structure, named \textit{RexNet}, is mainly composed by two parts, the \textit{REgionNet} and the \textit{conteXtNet}.
First, the \textit{RegionNet} is inspired by the Fast R-CNN framework~\cite{girshick2015fast}. Fast R-CNN is recently proposed for object detection and achieves superior performance because the convolutional features of entire image are shared and features of each patch (or RoI) are extracted via the RoI pooling layer. We extend Fast R-CNN to salient object detection by introducing mask-based RoI pooling and formulating salient object detection as a binary region classification task.
The image is first segmented into regions and are used as input of \textit{RegionNet}, the \textit{RegionNet} then predicts saliency score of each region end-to-end to form saliency map of the entire image. Since the regions are segmented by edge-preserved methods, saliency map generated by our network is naturally with sharp boundaries.

Second, the \textit{ContextNet} aims to provide strongly reliable multi-scale contextual information. 
Different from most previous works which consider context by expanding region window at a certain layer, in this paper, we consider to model context via multiple spatial scales. This is based on the observation that different layers of CNN represent different levels of semantic~\cite{Zeiler2013Visualizing, Li2015Convergent}, considering context of different levels may be more sufficient.
We achieve this by taking advantages of dense image prediction. For all max-pooling layers of \textit{RegionNet}, we attach  multiple convolutional layers to predict saliency map of different levels. Then all levels of saliency map are fused with \textit{RegionNet} to generate the final saliency map. Our method generates saliency map with accurate location while keeping fine object boundaries.

Other than the effectiveness, our proposed frameworks is efficient, since we take advantages of regions by extending the efficient Fast R-CNN framework, which predicts saliency score of regions by only one forwarding. We also extend our method to RGB-D saliency by applying depth refinement. Experiments on 2 RGB-D benchmark datasets demonstrate that the proposed \textit{RexNet} outperforms other methods by a large margin.

The main contributions of this paper are three-fold. 
First, we proposed \textit{RegionNet} which generates saliency score of regions efficiently and preserves object boundaries. 
Second, multi-scale spatial context is considered and attached to \textit{RegionNet} to boost salient object detection performance. 
Third, we extend our method to RGB-D saliency datasets and use depth information to further refine saliency maps.

The rest of this paper is organized as follows. Section~\ref{sec:relwork} discusses related work. Section~\ref{sec:regionnet} and Section~\ref{sec:context} introduce the details of the proposed \textit{RegionNet} and \textit{ContextNet} correspondingly. Section~\ref{sec:training} describes the training details of the proposed network. Section~\ref{sec:rgbd} introduces our extension to RGB-D salient object detection. Section~\ref{sec:exper} shows the experimental results and comparison with state-of-the-art methods. And conclusion is made in Section~\ref{sec:conclusion}.

\section{Related Work}\label{sec:relwork}
In this section, we introduce traditional salient detection methods and the recent CNN based methods. In addition, we also introduce some related works that integrate multi-scale context information and some topics related to salient object detection.
\subsection{Traditional Methods}

Salient object detection was first exploited by Itti \textit{et.al.}~\cite{itti1998model}, and later attracted wide attention in the computer vision society. Traditional methods mostly rely on prior assumptions and most are un-supervised. Center-surround difference which assumes that salient regions differs from their surrounding regions is an important prior in early research. Itti \textit{et.al.}~\cite{itti1998model} first proposed center-surround difference at different scales to compute saliency. Liu \textit{et.al.}~\cite{liu2007learning} propose center-surround histogram which defines saliency as the difference between center region and its surrounding region. Li \textit{et.al.}~\cite{li2013contextual} propose cost-sensitive SVM to learn and discover salient regions that are different from their surrounding regions. 
These methods cannot provide sharp boundary for salient region because they are based on rectangle regions, which is only able to generate coarse and blurry boundary. 

While center-surround difference considers local contrast, it does not take into consideration of global contrast. Global contrast based methods are later proposed, \eg, 
Cheng \textit{et.al.}~\cite{cheng2011global} and Yan \textit{et.al.}~\cite{yan2013hierarchical}. In~\cite{cheng2011global}, image is first segmented into superpixels. Then saliency value of each region is defined as the contrast with all other regions. The contrast is weighted by spatial distance so that nearby regions have greater impact on it. To deal with objects with complex structures, Yan \textit{et.al.}~\cite{yan2013hierarchical} propose a hierarchical model which analyzes saliency cues from multiple scales based on local contrast and then infers the final saliency values of regions by optimizing them in a tree model. Following them, many methods utilizing bottom-up priors are proposed, readers are encouraged to find more details in a recent survey paper by Borji \textit{et.al.}~\cite{borji2014salient}.

\subsection{CNN based Methods}
Deep Convolutional Neural Network (CNN) has attracted a lot of attention for its outstanding performance in representing high-level semantic. 
%Recent advances in salient object detection have also exploited CNN to represent high-level semantic. 
Here, we mention are few representative work. 
These work can be divided into two categories according to their treatment of input images: region-based methods and pixel-based methods. Region-based methods formulate salient object detection as a region classification task, namely, extracting features of regions and predict their saliency score. While pixel-based methods directly predict saliency map pixels-to-pixels with CNN.

\textbf{Region-based methods.} Wang \textit{et.al.}~\cite{wang2015deep} propose to detect salient object by integrating both local estimation and global search with two trained networks DNN-L and DNN-G. Zhao \textit{et.al.} ~\cite{zhao2015saliency} consider global and local context by putting a global and a closer-focused superpixel-centered window to extract features of each superpixel, respectively, and then combine them to predict saliency score.  
Li \textit{et.al.}~\cite{li2015visual} propose multi-scale deep features by extracting features of each region at three scales and then fuse them to generate its saliency score. 
These works are region-based which focused on extracting features of regions and fuse larger scale of regions as context to predict saliency score of each region. These fusions are mostly applied at only one layer and does not achieve a optimal performance.
In addition, the networks extract features of one region for each forwarding which is very time-consuming.

\textbf{Pixel-based methods.} Recently, CNN has also been applied to pixels-to-pixels dense image prediction, such as semantic segmentation and saliency prediction.
 Long \textit{et.al.}~\cite{long2015fully} propose fully convolutional networks which is trained end-to-end and pixels-to-pixels by introducing fully convolutional layers and a skip architecture. 
 Chen \textit{et.al.}~\cite{chen2015disc} propose a coarse-to-fine manner in which the first CNN generates coarse map using the entire image as input and then the second CNN takes the coarse map and local patch as input to generate fine-grained saliency map. 
 Li \textit{et.al.}~\cite{li2016deepsaliency} propose a multi-task model based on fully convolutional network. 
 In~\cite{li2016deepsaliency}, saliency detection task is in conjunction with object segmentation task, which is helpful for perceiving objects. A Laplacian regularized regression is then applied to refine saliency map.
However, while end-to-end dense saliency prediction is efficient, the resulting saliency maps are coarse and with 
blurry object boundaries due to the presence of convolutional layers with large receptive fields and pooling layers.

\subsection{RGB-D Salient Object Detection}\label{subsec:rgbdsal}
RGB-D saliency is an emerging topic and most RGB-D saliency methods are based on fusing depth priors with RGB saliency priors.
Ju \textit{et.al.}~\cite{ju2014depth} propose RGB-D saliency method based on anisotropic center-surround difference, in which saliency is measured as how much it outstands from surroundings. Peng \textit{et.al.}~\cite{peng2014rgbd} propose depth saliency with multi-contextual contrast and then fuse it with appearance cues via a multi-stage model. 
Ren \textit{et.al.}~\cite{ren2015exploiting} propose normalized depth prior and global-context surface orientation prior based on depth information and then fuse them with RGB region contrast priors. Depth contrast may cause false positives in background region, to address it, in~\cite{Feng2016Local}, Feng \textit{et.al.} propose local background enclosure feature based on the observation that salient objects tend to be locally in front of surrounding regions.
To the best of our knowledge, existing RGB-D salient object detection are all using hand-crafted features and the performance is not optimized.

\subsection{Multi-scale Context}\label{subsec:mscontext}
Multi-scale context has been proved to be useful for image segmentation task~\cite{hariharan2015hypercolumns, zhao2015saliency, li2015visual, liu2016dhsnet}. Hariharan~\textit{et.al.}~\cite{hariharan2015hypercolumns} proposed hypercolumns for object segmentation and fine-grained localization, in which they defined “hypercolumn” at a given input location as the outputs of all layers at that location. Features of different layers are combined and then be used for classification. 
Zhao~\textit{et.al.}~\cite{zhao2015saliency} proposed multi-context network which extracts features of a given superpixel at global and local scale, and then predict saliency value of that superpixel.
Li~\textit{et.al.}~\cite{li2015visual} proposed to extract features at three scales: bounding box, neighbourhood rectangular and the entire image.
Liu~\textit{et.al.}~\cite{liu2016dhsnet} proposed to use recurrent convolutional layers~(RCLs)~\cite{liang2015recurrent} iteratively to integrate context information and to refine saliency maps. At each step, the RCL takes coarse saliency map from last step and feature map at lower layer as input to predict a finer saliency map. In this way, context information is integrated iteratively and the final saliency map is more accurate than that predicted from global context.

The proposed \textit{ContextNet} differs from those at two aspects. First, the \textit{ContextNet} is a holistically-nested architecture~\cite{xie2015holistically} which predicts saliency map at each branch and fuse them finally. Second, we propose \textit{EdgeLoss} as a supervision which makes the boundary of segmentation result more clear.

\begin{figure*}[!t] %\footnotesize
\begin{center}
\includegraphics[width=1\linewidth,trim = 0mm 0mm 0mm 0mm, clip]{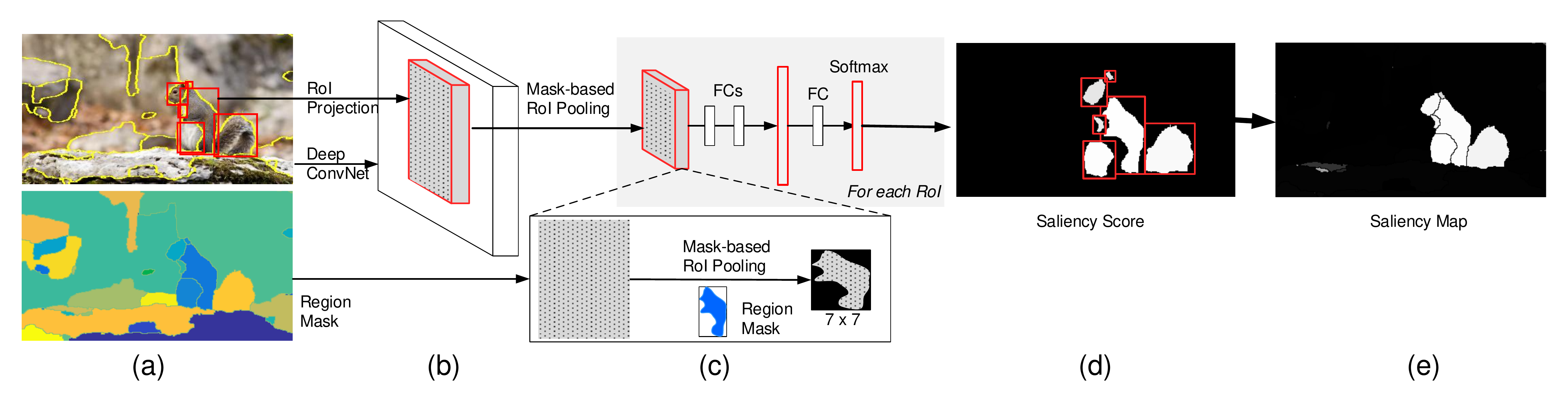}
\vspace{-8mm}
\caption{Pipeline of \textit{RegionNet}. We extend the Fast R-CNN framework for saliency detection. (a) Image is first segmented into regions and the region mask which records the index of regions is also generated. For each region, we use its external rectangle as RoI. Note that, for clarity, we only show RoIs of salient objects, the background regions are omitted. (b) All RoIs are put into the convolutional networks,  and (c) at the RoI pooling layer, the mask-based RoI pooling is applied to extract features inside region mask. In this way, the features of irregular region can be extracted. (d) With this mask-based pooling, the framework predicts saliency score of regions end-to-end, and (e) to form the saliency map of the entire image.}
\label{fig:edge_pre_rcnn}
\end{center}
\vspace{-2mm}
\end{figure*}

\subsection{Fixation prediction and semantic segmentation}\label{subsec:Fixation}
Fixation prediction~\cite{itti1998model,zhang2008sun,murray2011saliency,zhang2013saliency} aims to predict the regions people may pay attention to, and semantic segmentation~\cite{long2015fully, chen2014semantic} aims to segment objects of certain classes in images. They are topics related to salient object detection, but they also have significant differences. Fixation prediction aims to predict \textit{regions} which most attract people's attention, while salient object detection focuses on segmenting the most attractive \textit{objects}. For semantic segmentation, saliency detection is a class-agnostic task, whether an object is salient or not is largely depend on its surroundings, while semantic segmentation mainly focuses on segmentation objects of certain classes (\textit{e.g.} 20 classes in PASCAL VOC dataset). So compared with semantic segmentation, context information is more important for saliency detection, and this is the main motivation of our \textit{ContextNet}.

\section{RegionNet: Edge Preserving Neural Network for Salient Object Detection}\label{sec:regionnet}
\subsection{Motivation}
In this paper, we aim to propose a unified framework which can preserve object boundaries and take multi-scale spatial context into consideration. To preserve object boundaries, we propose an effective network, named \textit{RegionNet}, which generates saliency score of each region end-to-end (\fref{fig:edge_pre_rcnn}). Different from previous region-based methods~\cite{wang2015deep, zhao2015saliency, li2015visual}, we extend the efficient Fast R-CNN framework~\cite{girshick2015fast} for salient object detection for the first time.
On the other hand, previous works consider context mainly by expanding window of region or using entire images at a certain data or feature layer. In this paper, we consider context at multiple layers and using dense saliency prediction framework to generate saliency maps to complement \textit{RegionNet}. The architecture of the proposed framework is shown in \fref{fig:framework}.

In this section, we first introduce the idea of edge-preserving saliency detection based on a CNN network. This idea is previously appeared in our conference paper~\cite{fl2016}. In section~\ref{sec:context}, we extend this idea with consideration of multi-scale spatial context.

\begin{figure*}[!t] %\footnotesize
\begin{center}
\includegraphics[width=1\linewidth,trim = 0mm 0mm 0mm 0mm, clip]{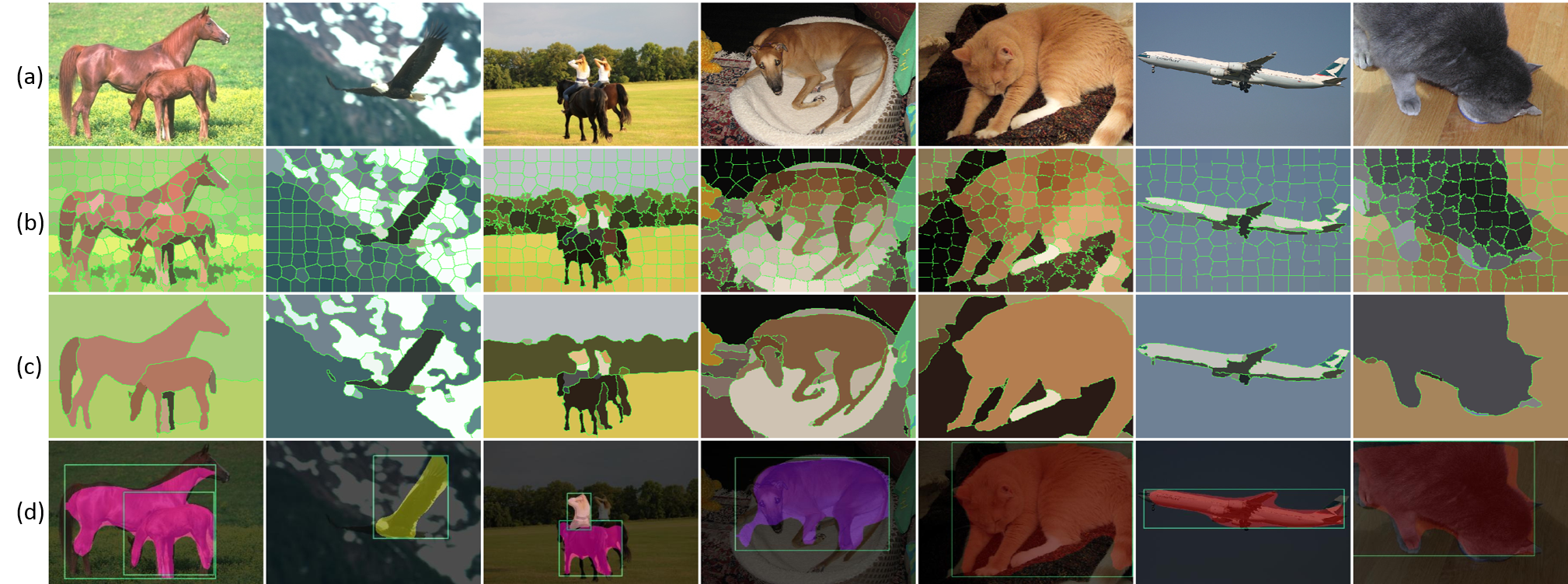}
\vspace{-4mm}
\caption{(a) images, (b) and (c) superpixel regions and edge regions. Pixels in each region are replaced with their mean color, (d) masks generated by MNC~\cite{dai2016instance}. (i) We can see that edges divide images into fewer regions than superpixels and thus preserving more compactness of objects, which is helpful for saliency prediction. (ii) The superpixels and edges regions achieve higher boundary accuracy than masks generated by MNC~\cite{dai2016instance}. Best viewed in color.}
\label{fig:compare_regions}
\end{center}
\vspace{0mm}
\end{figure*}

\subsection{RegionNet}
In this section, we introduce \textit{RegionNet} which takes advantage of CNN for high effectiveness and high efficiency. More importantly, it takes advantage of region segmentation which enables clear detection boundary and further improves the accuracy.

\noindent\textbf{Network architecture. }
We extend original Fast R-CNN ~\cite{girshick2015fast} structure for end-to-end saliency detection. 
Fast R-CNN is an efficient and general framework in which the convolutional layers are shared on the entire image and the feature of each region is extracted by the RoI pooling layer. However, to the best of our knowledge, Fast R-CNN is only used for object detection and classification but not for saliency. Namely, the result of Fast R-CNN is bounding box but not pixel-wise. In this paper, we make the modification to enable edge preserving saliency by introducing mask-based RoI pooling. Different from previous region-based methods which deal with each region of an image independently, our proposed Fast R-CNN structure processes all regions end-to-end and with the entire image considered.

\noindent\textbf{Detection pipeline. }
As illustrated in \fref{fig:edge_pre_rcnn}, first, given an image, we segment it into regions using superpixel and edges. And for each region, we use its external rectangle as proposal (or RoI) and use it as input of Fast R-CNN framework similar with object detection tasks. We also generate a region mask with the same size of image to record the region index for each pixel and then downsample it by 16 times and put it into the RoI pooling layer.

Then, at the RoI pooling stage, features inside each RoI ($h\times w$) are pooled into a fixed scale $H\times W$ ($7\times 7$ in our work). So each sub-window with scale $h/H \times w/W$ is converted to one value with max-pooling. To extract feature of irregular pixel-wise RoI region, we only pool features inside its region mask while leaving others as $0$.
The process of the proposed mask-based RoI pooling is formulated as following. For region with index $i$, and a certain sub-window as $SW_j$, we denote region mask as $M$, features before pooling as $F$, the pooled feature at sub-window $SW_j$ as $P_j$, then
\begin{equation}
P_j=\left\{
\begin{array}{rcl}
\max\limits_{\{k|k\in{SW_j}, {M_k = i}\}}{F_k}       &      & i \in M(SW_j),\\
0     &      & i \notin M(SW_j).
\end{array} \right.
\end{equation} 
With this mask-based pooling, features of each region are extracted and the edge information is also preserved. 	

Last, by considering salient object detection as a binary classification problem, the network generates saliency score of regions to form the saliency map of entire image end-to-end.

Note that, in our work, to segment image into regions, besides superpixel, we also consider larger scale regions which are segmented by edges (denoted as edge regions). This is based on the observation that when an object is segmented into dozens of superpixels, it will be difficult to uniformly highlight the whole object. The edge regions can preserve more compactness of objects and thus may be more effective. Recent advances in edge detection have achieved highly satisfactory performance which makes it practical to use edge information to help better detect salient objects. In our work, we use HED method of Xie \textit{et.al.}~\cite{xie2015holistically} to get object edges and then thinning them using method of Dollar \textit{et.al.}~\cite{dollar2013structured}. The superpixel is segmented using SLIC algorithm~\cite{achanta2012slic}.

Some examples of superpixel regions and edge regions are shown in \fref{fig:compare_regions}. We can see that edges segment image into fewer regions and better preserves compactness of object. For region-based methods, this will help improve the final performance and since the number of regions is smaller, it also reduces computation cost. Considering the fault-tolerant capability, namely, misclassification of edge regions may decrease performance largely, the superpixel regions are also used in our method. These two scales regions are complementary since superpixel regions can generate results with high resolution and edge regions can preserve more compactness of objects. 

Note that the similar idea of mask-based RoI pooling has also been applied in MNC~\cite{dai2016instance} for semantic segmentation. However, we have much difference. In~\cite{dai2016instance}, the masks were generated by the multi-task network and they are continuous values in $[0, 1]$. The masked feature is the element-wise product of features and masks. While in our work, the masks are got by segmenting images into regions with superpixels~\cite{achanta2012slic} and edges~\cite{xie2015holistically}, they are binary and the mask-based RoI pooling is to extract features inside the masks. %Compared with masks in~\cite{dai2016instance}, our region masks have higher boundary accuracy.
The SLIC algorithm~\cite{achanta2012slic} for generating superpixels has strong ability to adhere to image boundaries, so its boundary accuracy is quite good. The HED~\cite{xie2015holistically} network is designed for edge detection, the boundary accuracy is much better than multi-task networks in~\cite{dai2016instance}. So the masks of our method has higher boundary accuracy compared with MNC~\cite{dai2016instance}. Some examples are shown in \fref{fig:compare_regions}.

We denote the saliency map generated by \textit{RegionNet} with superpixel regions and edge regions as $S_S$ and $S_E$, respectively. We have shown in our previous conference paper~\cite{fl2016} that $S_E$ outperforms most previous works, and the combination of $S_E$ and $S_S$ achieves better performance, which shows the effectiveness of edge regions and the combination with superpixel regions. More detailed experimental results are shown in Section~\ref{sec:exper}.

\begin{figure*}[!t] %\footnotesize
\begin{center}
%\begin{tabular}{cccc}
\begin{tabular}{c} % 42 40 28
%\hspace{-3mm}\rotatebox{90}{\hspace{38mm}\textbf{ECSSD}}  &\hspace{-0.8cm}\hspace{2mm}
%\hspace{2mm}
\includegraphics[width=0.9\linewidth,trim = 0mm 0mm 0mm 0mm, clip]{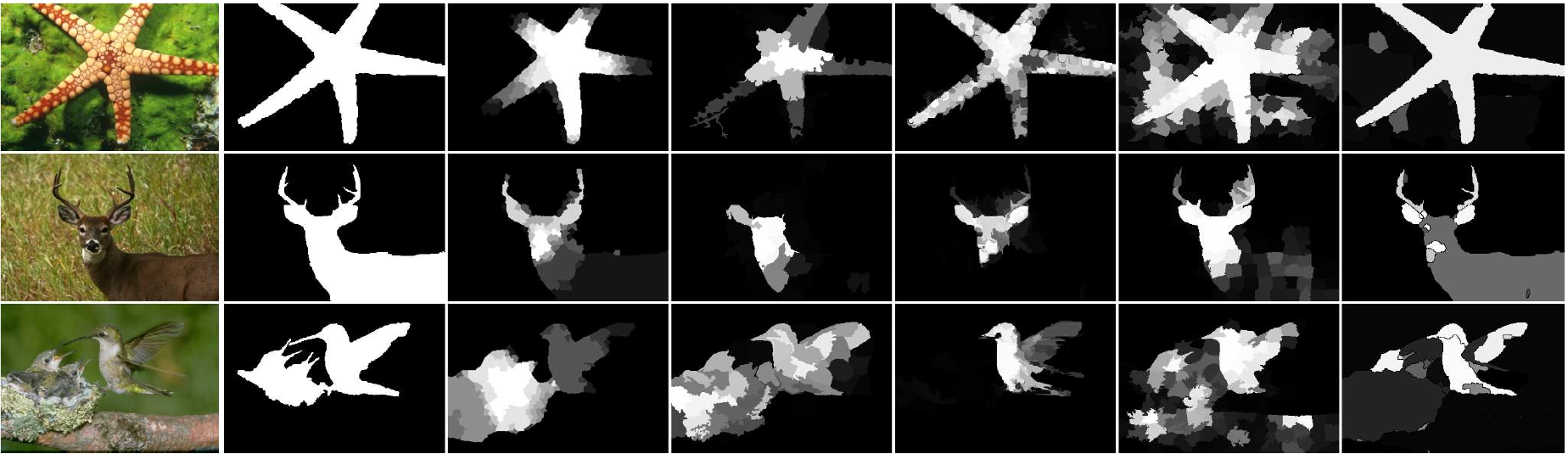}\\

\begin{minipage}[b]{0.05\linewidth}
  \centering
  \centerline{\footnotesize{}}\medskip
\end{minipage}
\begin{minipage}[b]{0.12\linewidth}
  \centering
  \centerline{\footnotesize{Image}}\medskip
\end{minipage}
\begin{minipage}[b]{0.12\linewidth}
  \centering
  \centerline{\footnotesize{GT}}\medskip
\end{minipage}
\begin{minipage}[b]{0.12\linewidth}
  \centering
  \centerline{\footnotesize{LEGS~\cite{wang2015deep}}}\medskip
\end{minipage}
\begin{minipage}[b]{0.12\linewidth}
  \centering
  \centerline{\footnotesize{MC~\cite{zhao2015saliency}}}\medskip
\end{minipage}
\begin{minipage}[b]{0.12\linewidth}
  \centering
  \centerline{\footnotesize{MDF~\cite{li2015visual}}}\medskip
\end{minipage}
\begin{minipage}[b]{0.12\linewidth}
  \centering
  \centerline{\footnotesize{Ours ($S_S$)}}\medskip
\end{minipage}
\begin{minipage}[b]{0.12\linewidth}
  \centering
  \centerline{\footnotesize{Ours ($S_E$)}}\medskip
\end{minipage}
\begin{minipage}[b]{0.05\linewidth}
  \centering
  \centerline{\footnotesize{}}\medskip
\end{minipage}
\\
\hfill
\vspace{-5 mm}
\end{tabular}

\caption{Results of previous region-based methods and our $S_S$ and $S_E$. We can see that misclassification of regions has a great impact on the final performance and most regions are assigned to near either 0 or 1, with few intermediate values. These will limit the precision at high recall when thresholding.}
\label{fig:compare_regSalMap}
\end{center}
\vspace{-2mm}
\end{figure*}

\begin{figure*}[!t] %\footnotesize
\begin{center}
%\begin{tabular}{cccc}
\begin{tabular}{ccc} % 42 40 28
%\hspace{-3mm}\rotatebox{90}{\hspace{38mm}\textbf{ECSSD}}  &\hspace{-0.8cm}\hspace{2mm}
%\hspace{2mm}
\includegraphics[width=0.133\linewidth,trim = 0mm 0mm 0mm 0mm, clip]{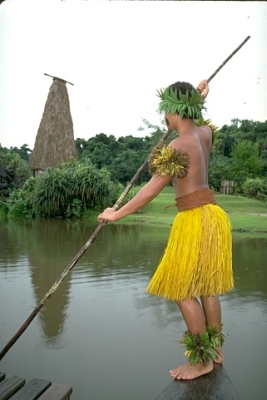}\hspace{0mm}
\includegraphics[width=0.2\linewidth,trim = 0mm 0mm 0mm 0mm, clip]{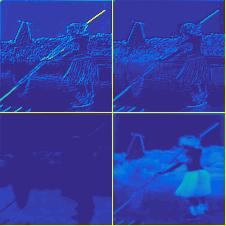}\hspace{0mm}
\includegraphics[width=0.2\linewidth,trim = 0mm 0mm 0mm 0mm, clip]{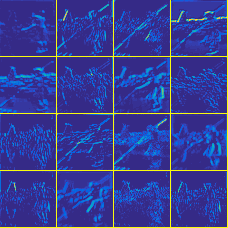}\hspace{0mm}
\includegraphics[width=0.2\linewidth,trim = 0mm 0mm 0mm 0mm, clip]{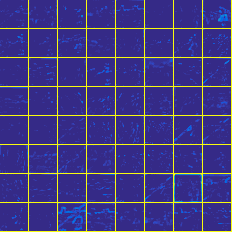}\hspace{0mm}
\includegraphics[width=0.2\linewidth,trim = 0mm 0mm 0mm 0mm, clip]{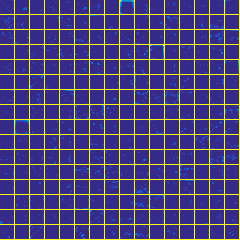}\hspace{0mm} \vspace{0 mm} \\

\begin{minipage}[b]{0.133\linewidth}
  \centering
  \centerline{\footnotesize{Image}}\medskip
\end{minipage}
\begin{minipage}[b]{0.2\linewidth}
  \centering
  \centerline{\footnotesize{pool1}}\medskip
\end{minipage}
\begin{minipage}[b]{0.2\linewidth}
  \centering
  \centerline{\footnotesize{pool2}}\medskip
\end{minipage}
\begin{minipage}[b]{0.2\linewidth}
  \centering
  \centerline{\footnotesize{pool3}}\medskip
\end{minipage}
\begin{minipage}[b]{0.2\linewidth}
  \centering
  \centerline{\footnotesize{pool4}}\medskip
\end{minipage}\\
\hfill
\vspace{-8 mm}
\end{tabular}

\caption{Visualization of features in different layers of \textit{RegionNet}. For a test image, we forward it in our trained \textit{RegionNet}, and then we extract features of the first four pooling layers and show each channel of them. Different layer represents different level of semantic. Best viewed in color.}
\label{fig:vis_feature}
\end{center}
\vspace{0mm}
\end{figure*}

\section{ContextNet: Multi-scale Contextual Neural Network for Salient Object Detection}\label{sec:context}

In this section, we introduce the extension of the proposed method by utilizing multi-scale context. In Section~\ref{subsec:motivation}, we first introduce the motivation for multi-scale context, after that, in Section~\ref{subsec:Architecture}, we introduce the architecture of the proposed multi-scale contextual network. In Section~\ref{subsec:Loss}, we introduce the loss function for supervising the \textit{ContextNet}, and in Section~\ref{subsec:Supervision}, we introduce deep supervision to accelerate convergence and improve prediction performance.

\subsection{Motivation}\label{subsec:motivation}

Salient object detection is a class-agnostic task, whether a region is salient or not is largely depend on its surroundings, \textit{i.e.}, context. While the \textit{RegionNet} we proposed can generate saliency map with well preserved boundary, it lacks of context information. In addition, region-based CNN methods~\cite{wang2015deep,zhao2015saliency,li2015visual} suffer from some common drawbacks. First, region-based methods are based on binary region classification, misclassification of regions will cause large false detection. Second, solving binary classification problem with huge amount of images using CNN causes the classification results to be extremely separated to either 0 or 1, thus saliency map is not smooth. These two issues will limit the precision at high recall. \fref{fig:compare_regSalMap} shows some results of previous region-based CNN methods and our $S_S$ and $S_E$.

As explored in previous works~\cite{Zeiler2013Visualizing, Li2015Convergent}, features in different layers of CNN has different properties and represent different levels of semantic. So fusing context from multiple layers may be more sufficient. \fref{fig:vis_feature} shows the visualization example of features in the first four pooling layers of \textit{RegionNet}. We can see that shallow layers mainly focus on bottom features, such as contour, and deep layers focus on more abstract high-level features. Based on these observations, in this paper, we consider context information by introducing multi-scale contextual layers, named \textit{ContextNet}, to address the issues mentioned above and to complement \textit{RegionNet}.

\begin{figure*}[!t] %\footnotesize
\begin{center}
\includegraphics[width=0.9\linewidth,trim = 0mm 0mm 0mm 0mm, clip]{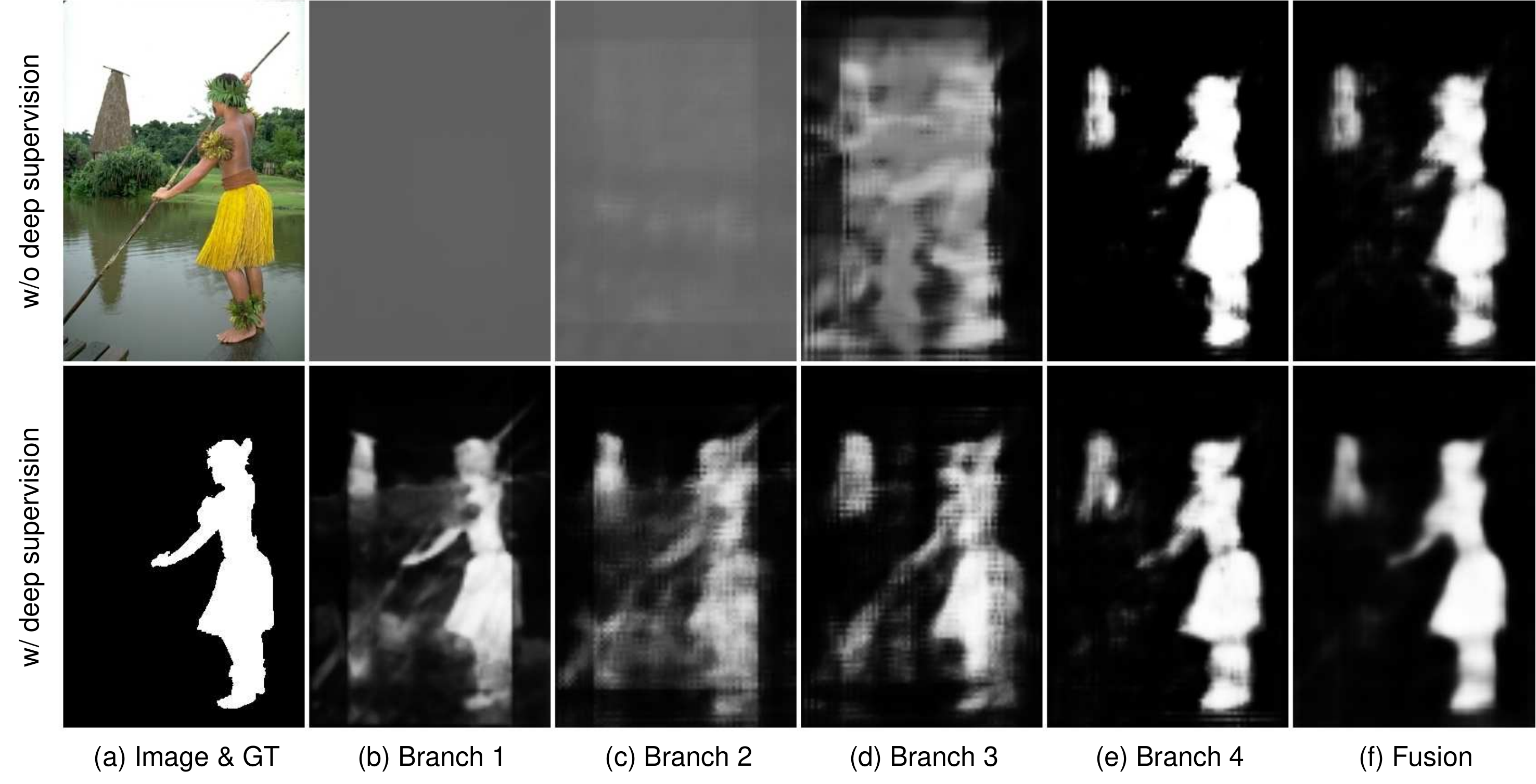}
\vspace{-2mm}
\caption{Effect of deep supervision. From left to right are image and ground truth, results of 4 branches, and fusion of all branches. The first row shows results without deep supervision and the second row shows results with deep supervision. Without deep supervision, the first and second branch learn almost nothing in our network due to the heavy bias.}
\label{fig:deep_supervision}
\end{center}
\vspace{0mm}
\end{figure*}

\subsection{Network Architecture}\label{subsec:Architecture}
The architecture of our proposed network is shown in \fref{fig:framework}. Based on the \textit{RegionNet}, we propose to use multi-scale dense image prediction method to model the relationship between regions and the global scenes at multiple levels. For all max pooling layers (except the RoI pooling layer) of \textit{RegionNet}, we attach five convolutional layers (called as branch) to predict saliency maps of different levels. The first three layers of each branch are with $3\times 3$ convolutional filters and 64, 64, 128 channels, and the dilated convolution~\cite{chen2014semantic} is also applied to increase the receptive field. The last two layers are fully convolutional layers with 128 and 1 channels.

Experimental results in~\cite{long2015fully} have demonstrated that denser prediction map has better performance. Following that, we propose to generate saliency map with one eighth scale of the original input images. So we set the stride of each branch as 4, 2, 1, 1, respectively. Note that the last branch is connected to the convolution layer before the fourth max-pooling layer, \textit{i.e.,} conv4\_3 in VGG16~\cite{vgg16}, so output of all branches have the same dimensions. The outputs of all branches are then fed into fully convolutional layers which learn the combination weights to generate saliency map $S_C$. The final saliency map $S$ is then got by fusing $S_S$, $S_E$, and $S_C$ via a fully convolutional layer.
\begin{equation}
S = Fusion(S_S, S_E, S_C).
\end{equation}

\subsection{Loss}\label{subsec:Loss}
We assume that the training data, $\mathcal{D} = \{(X_i, T_i)\}_{i=1}^N$, consists of $N$ training images and groundtruth. Our goal is to train a convolutional network $f(X; \theta)$ to predict saliency map of a given image. We define two kinds of loss for \textit{ContextNet} to generate saliency map with high accuracy and clear object boundary. 

The first \textit{Loss} is common used Cross Entropy Loss~$\mathcal{L_C}$, which aims to make the output saliency map $f(X; \theta)$ consistent with the groundtruth $T$.
\begin{equation}
\mathcal{L_C} = -\frac{1}{N}\sum\limits_{i=1}^{N}[T_ilog(f(X_i;\theta)) + (1-T_i)log(1-f(X_i;\theta))]
\end{equation}

The second \textit{Loss} is Edge Loss~$\mathcal{L_E}$ which aims to preserve edge and make the saliency map more uniform. Since we have segmented image into regions with edge-preserved methods, our assumption is that saliency map in the same region should share similar value, so that the final saliency map can also preserve edge and be more uniform. We average saliency map $f(X; \theta)$ in each region and marked the averaged map as \textbf{$\bar{f}(X; \theta)$}. The Edge Loss is defined as the $L2$ norm between saliency map $f(X; \theta)$ and the averaged map \textbf{$\bar{f}(X; \theta)$}.
\begin{equation}
\mathcal{L_E} = \frac{1}{2N}\sum\limits_{i=1}^{N}\|f(X_i; \theta) - \bar{f}(X_i; \theta)\|_2^2
\end{equation}

\subsection{Deep Supervision}\label{subsec:Supervision}

The proposed \textit{ContextNet} comprises of a fusion layer which fuses the outputs of four branches. Supervision only in the last fusion layer may cause heavy bias, namely, some layers may not be optimized adequately. To address this issue, in this paper, we utilize deep supervision~\cite{lee2015deeply,xie2015holistically} method, namely, outputs of all branches and their fusion result are also supervised.
\fref{fig:deep_supervision} shows the comparison of results with and without deep supervision. Without deep supervision, the network will be heavily biased towards some maps, and in extreme cases, some branches will learn nothing, \textit{e.g.}, \fref{fig:deep_supervision} (b) and (c). While with deep supervision, each branch learns and predicts saliency map with features at different scale, which accelerates convergence of the network and makes the final saliency map more precise.

\begin{figure*}[!t] %\footnotesize
\begin{center}
\includegraphics[width=0.9\linewidth,trim = 0mm 0mm 0mm 0mm, clip]{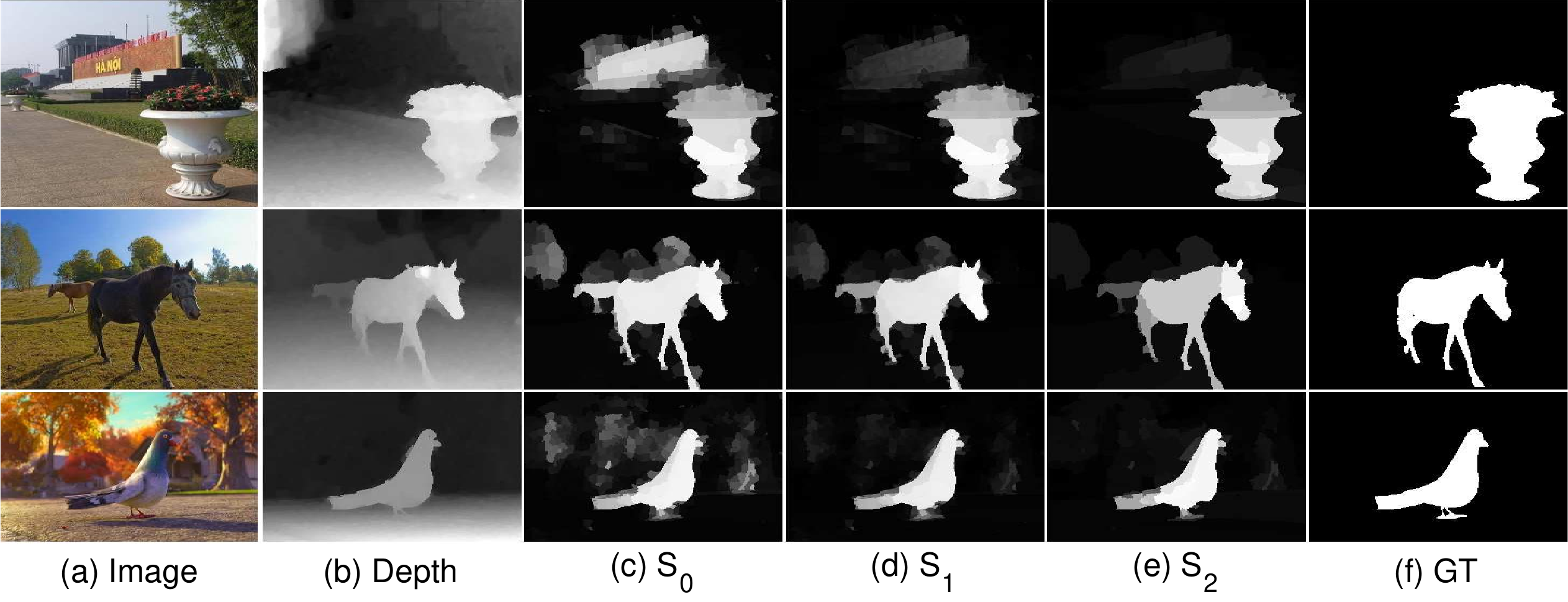}
\vspace{-2mm}
\caption{The process of depth refinement. (a) Image, (b) depth, (c) saliency map of our method using RGB data ($S_0$), (d) with the position prior, the background noise is strongly suppressed ($S_1$), and (e) with the local compactness prior, the background is further suppressed and the result map is more uniform ($S_2$), (f) groundtruth.}
\label{fig:depth_refinement}
\end{center}
\vspace{0mm}
\end{figure*}

\section{Network Training}\label{sec:training}
We implement our method using Caffe framework~\cite{jia2014caffe}. The training process consists of two stages. At the first stage, we fine-tune the \textit{RegionNet} using weights pre-trained on ImageNet~\cite{russakovsky2015imagenet}. At the second stage, we fix the weights of \textit{RegionNet} and then optimize the weights of the \textit{ContextNet} using SGD procedure. 

For the training of \textit{RegionNet}, a region is considered as salient/background if more than $80\%$ of its pixels are located inside/outside ground truth. The \textit{RegionNet} formulates salient object detection as a binary classification problem and the loss function we used is softmax loss. 
Following previous works, we fine-tune our \textit{RegionNet} based on VGG16~\cite{vgg16} which is pre-trained on ImageNet~\cite{russakovsky2015imagenet}.

For the training of \textit{ContextNet}, deep supervision is applied to accelerate convergence and to improve the final performance.

\section{Extension to RGB-D Salient Object Detection}\label{sec:rgbd}

Depth information is an important cue for salient object detection, especially for images with complex scenes. In this paper, we apply depth information to further improve the performance by extending our framework to RGB-D saliency datasets.

For RGB-D datasets, a simple idea is to train our network using RGB-D data directly. However, it suffers from two problems. First, our network is pre-trained on ImageNet~\cite{russakovsky2015imagenet}, it is unreasonable to fine-tune it using RGB-D data. Second, the image number of existing RGB-D saliency dataset is too small to well train a network. So in this paper, we propose to first generate saliency map using RGB data, and then refine it with depth information.

We propose two efficiency priors based on our observations: position prior and local compactness prior. For position prior, in most scenes, the salient object is located at the most front position. For local compactness prior, regions with similar depth, appearance and position should share similar saliency value.

We denote saliency map generated by our network as $S_0$. For position prior, we directly multiply $S_0$ by depth $D$ using a sigmoid function and denote it as $S_1$,
\begin{equation}
S_1 = S_0 \times \frac{1}{1 + exp(-\sigma \times D)},
\end{equation}
in which the parameter $\sigma$ is set to 5 empirically in our work.
Note that we have transformed the depth similar with~\cite{ju2014depth}, in which the depth is rescaled to~$[0, 1]$ and pixels with shorter distance are attached with larger intensity.

For local compactness prior, saliency value of each region $S_2(i)$ is refined with their neighbor regions $\mathcal{N}(i)$ weighted by depth and appearance similarity.
\begin{equation}
S_2(i) = \sum\limits_{j\in{ \mathcal{N}(i)}}W(i,j)S_1(j),
\end{equation}
with
\begin{equation}
W(i,j) = exp(-\frac{D(i,j)^2}{2\sigma_{dep}^2})exp(-\frac{Col(i,j)^2}{2\sigma_{col}^2}),
\end{equation}
in which $Col(i,j)$ denotes the Euclidean distance of RGB color. We set $\sigma_{dep} = 0.02$ and $\sigma_{col} = 5$ empirically in our work. \fref{fig:depth_refinement} shows some examples of the depth refinement.

\section{Experiments}\label{sec:exper}

To evaluate the effectiveness of each component and study the performance of the proposed method, we conduct experiments on six RGB and two RGB-D benchmark datasets and compare our method with state-of-the-art methods quantitatively and qualitatively.

\begin{figure*}[!t] %\footnotesize
\begin{center}
%\begin{tabular}{cccc}
\begin{tabular}{ccc} % 42 40 28
%\hspace{-3mm}\rotatebox{90}{\hspace{38mm}\textbf{ECSSD}}  &\hspace{-0.8cm}\hspace{2mm}
\hspace{2mm}
\includegraphics[width=0.33\linewidth,trim = 0mm 0mm 0mm 0mm, clip]{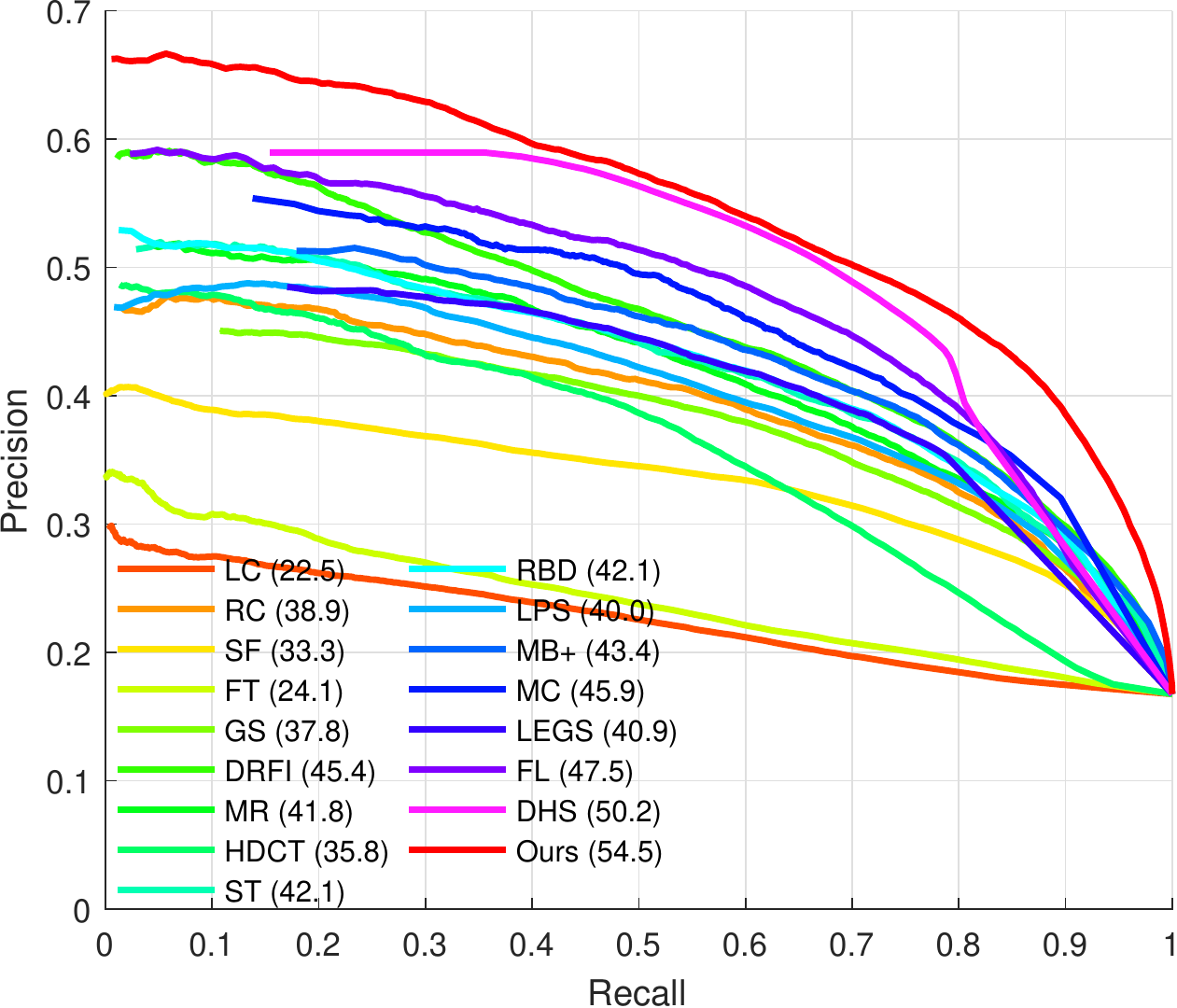}\hspace{3mm}
\includegraphics[width=0.33\linewidth,trim = 0mm 0mm 0mm 0mm, clip]{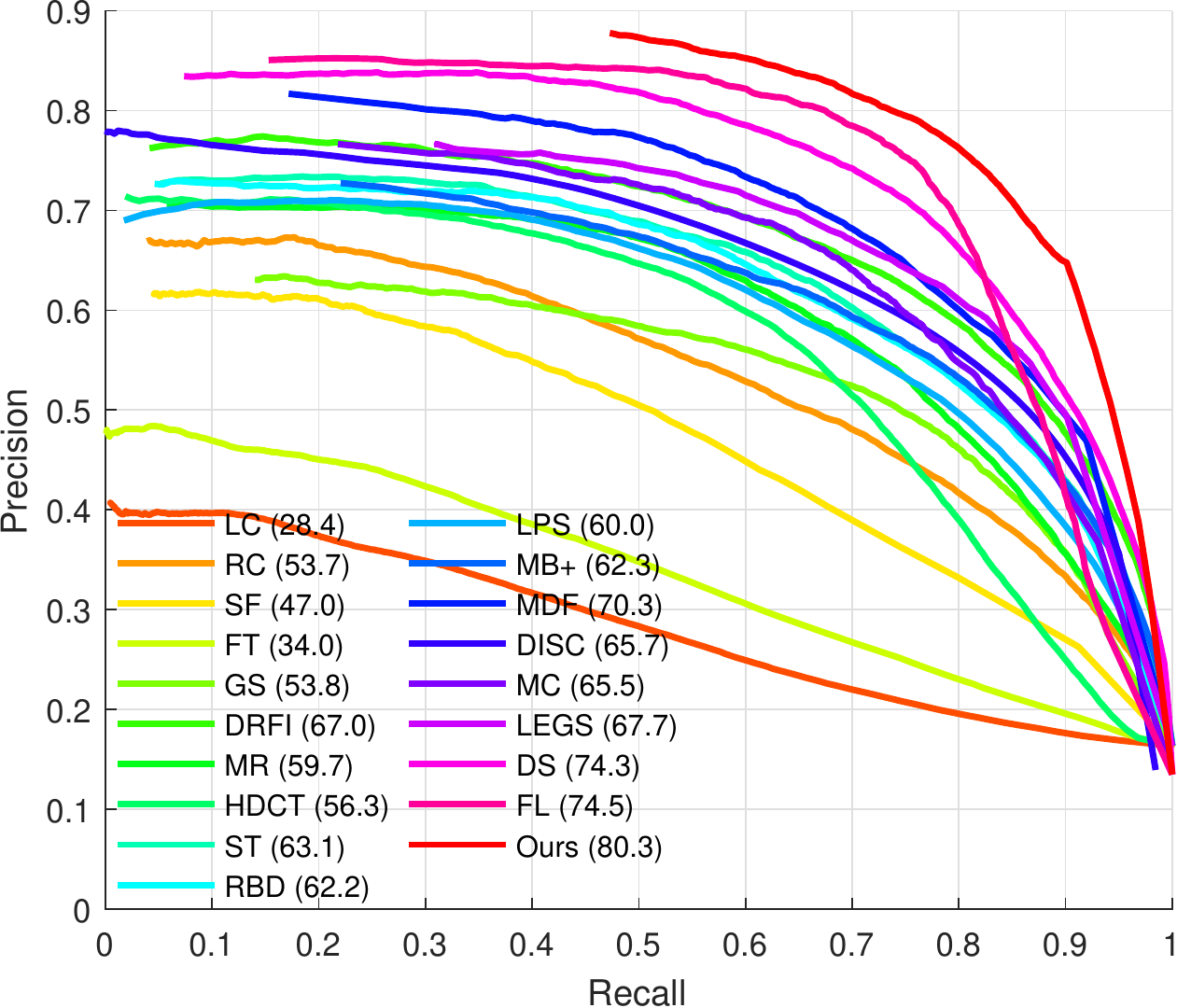}\hspace{3mm}
\includegraphics[width=0.33\linewidth,trim = 0mm 0mm 0mm 0mm, clip]{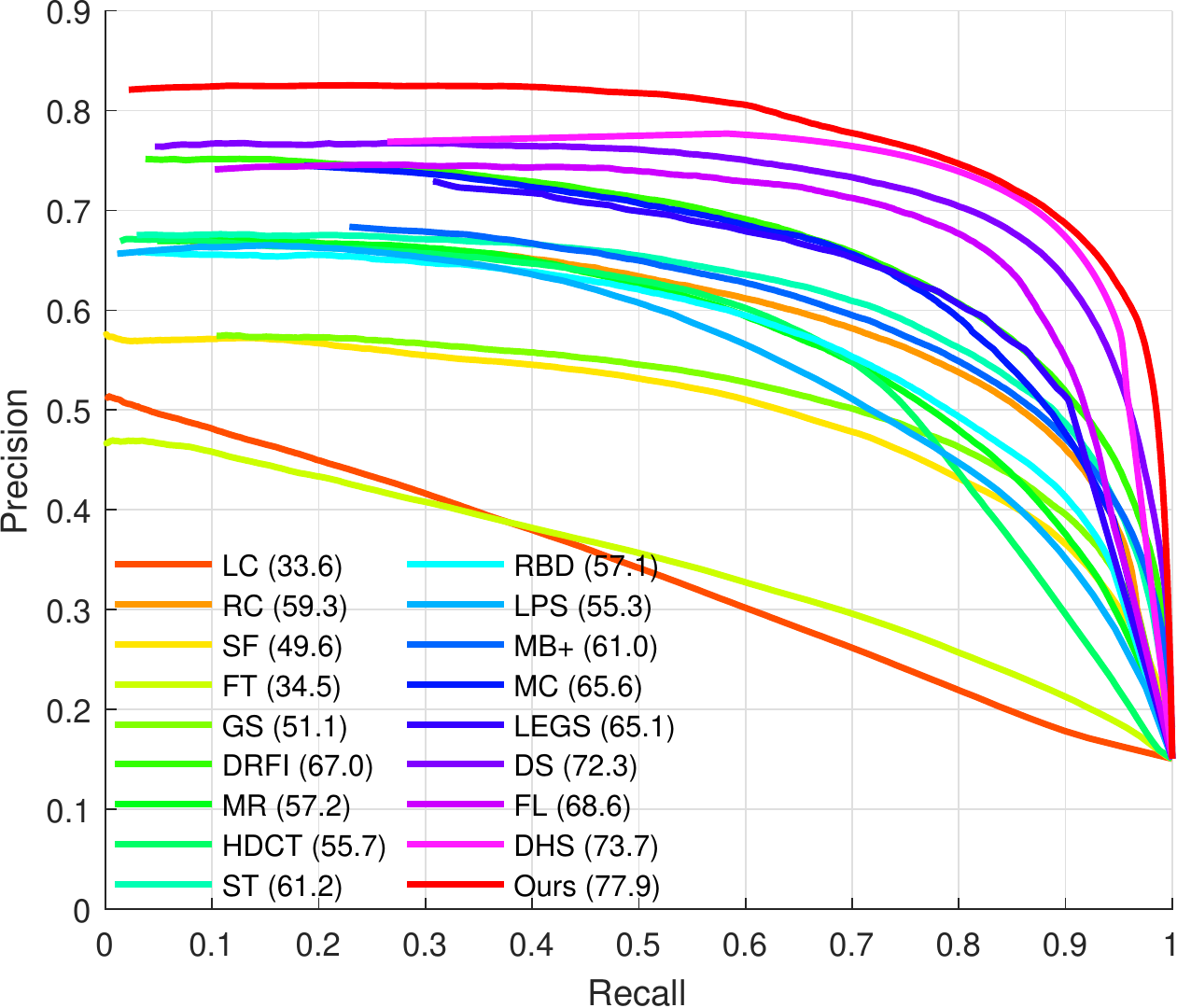}\hspace{2mm} \vspace{1 mm} \\

%\hspace{-3mm}\rotatebox{90}{\hspace{26mm}\textbf{DUT-ORMON}}  &\hspace{-0.8cm}
\includegraphics[width=0.33\linewidth,trim = 0mm 0mm 0mm 0mm, clip]{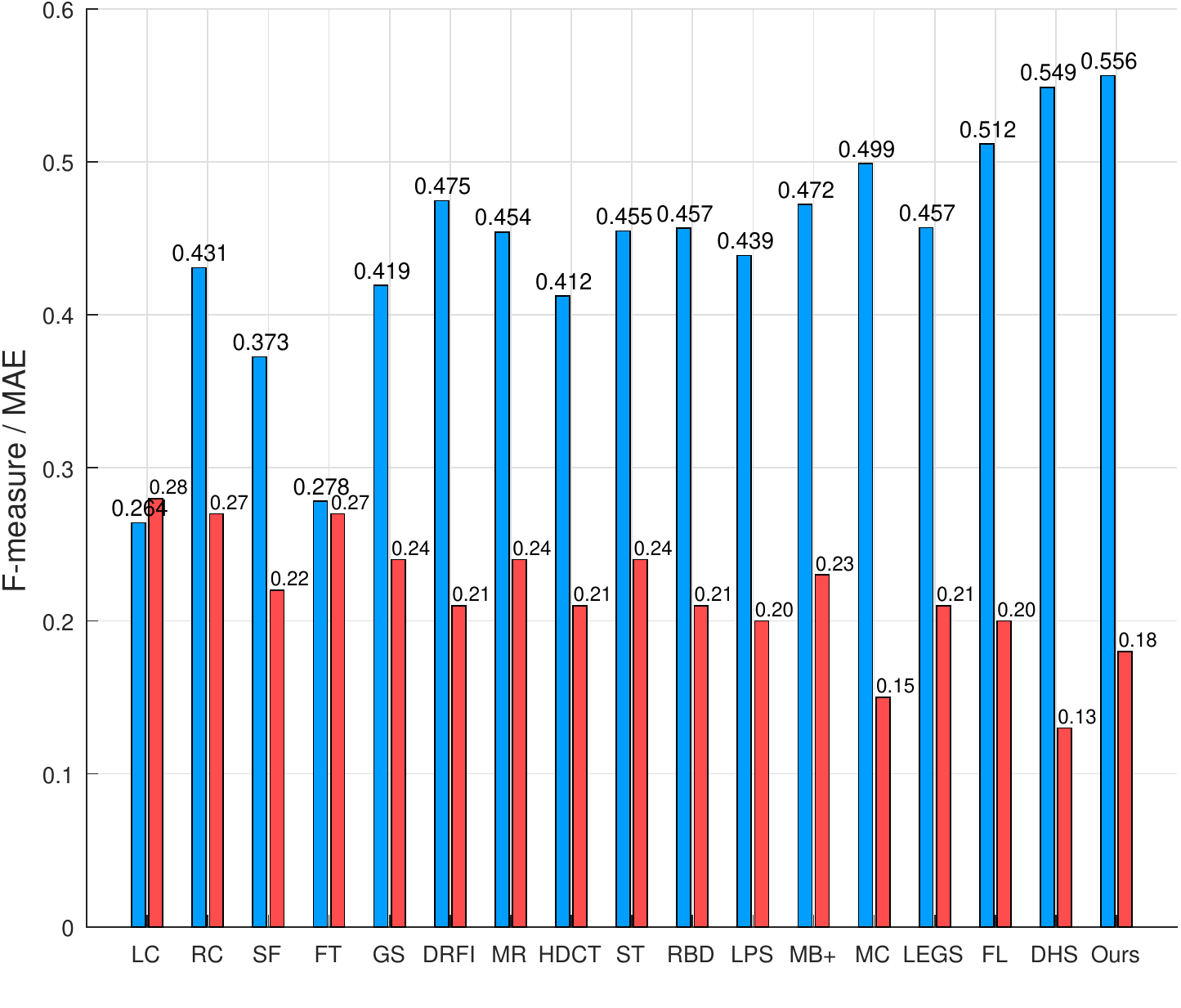}\hspace{3mm}
\includegraphics[width=0.33\linewidth,trim = 0mm 0mm 0mm 0mm, clip]{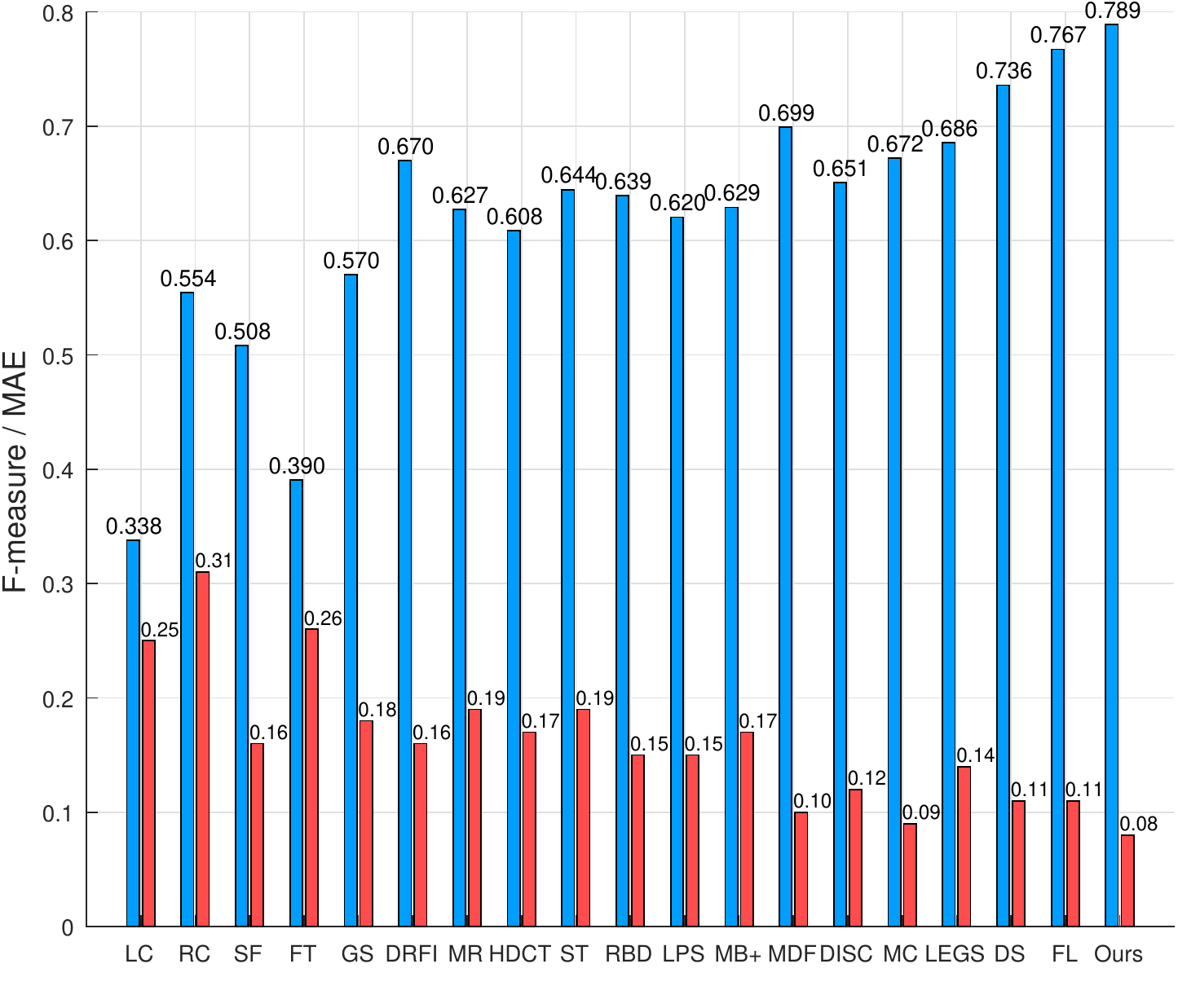}\hspace{3mm}
\includegraphics[width=0.33\linewidth,trim = 0mm 0mm 0mm 0mm, clip]{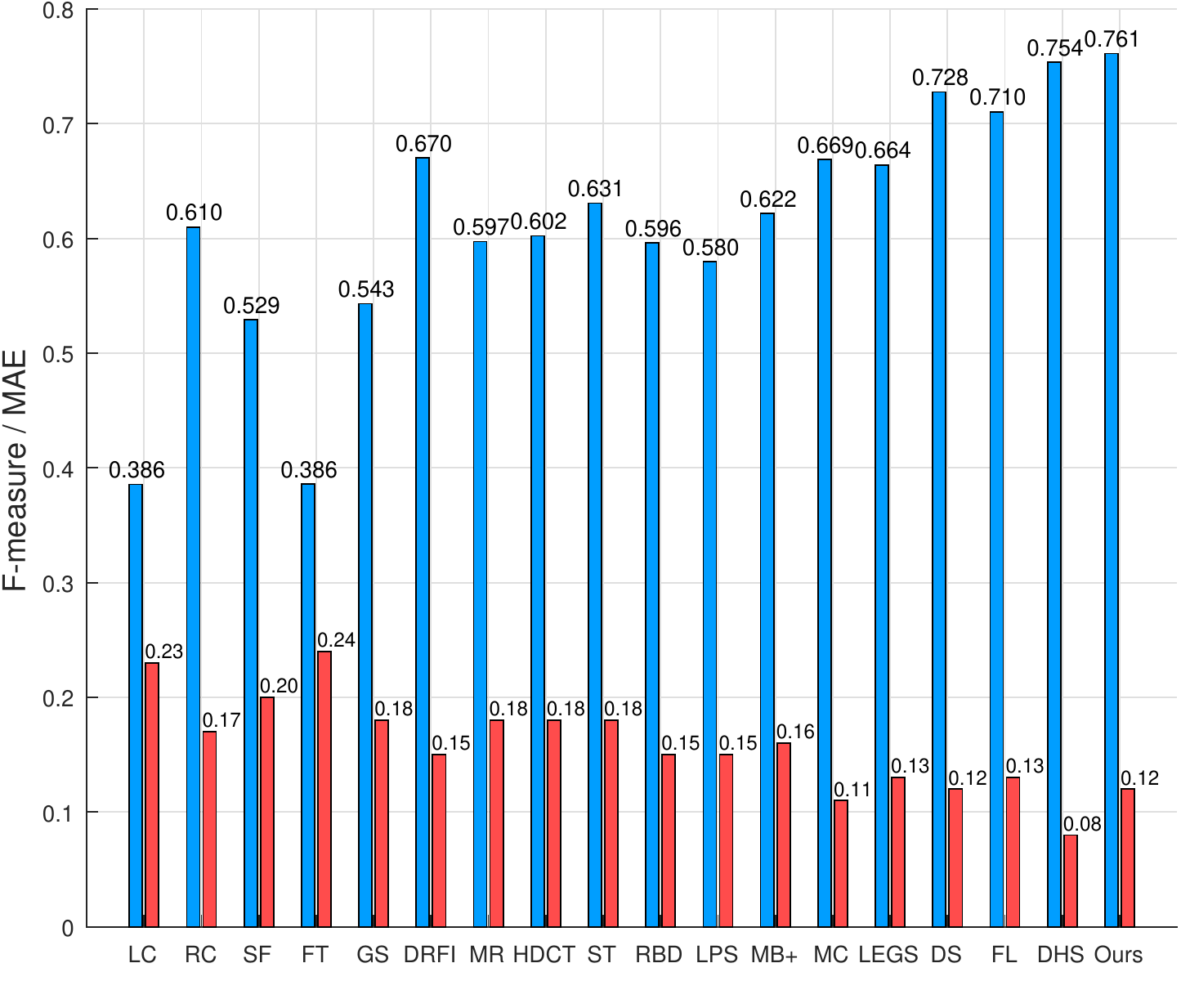}\hspace{2mm}\vspace{1 mm}\\

\begin{minipage}[b]{0.33\linewidth}
  \centering
  \centerline{\footnotesize{\textbf{JuddDB}}}\medskip
\end{minipage}
\begin{minipage}[b]{0.33\linewidth}
  \centering
  \centerline{\footnotesize{\textbf{DUT-OMRON}}}\medskip
\end{minipage}
\begin{minipage}[b]{0.33\linewidth}
  \centering
  \centerline{\footnotesize{\textbf{THUR15K}}}\medskip
\end{minipage}\\
\hfill
\vspace{-2 mm}\\
\hspace{2mm}
\includegraphics[width=0.33\linewidth,trim = 0mm 0mm 0mm 0mm, clip]{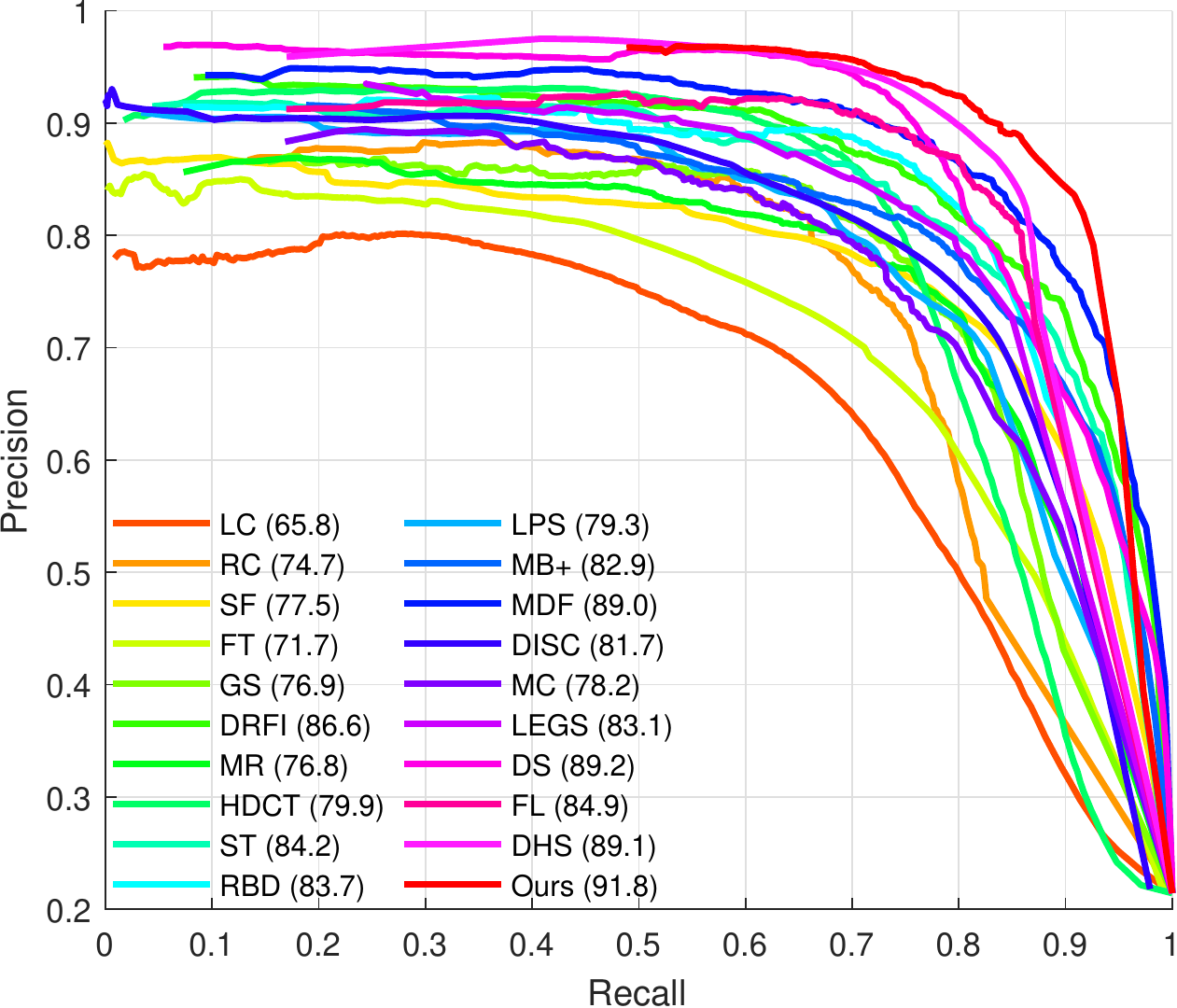}\hspace{3mm}
\includegraphics[width=0.33\linewidth,trim = 0mm 0mm 0mm 0mm, clip]{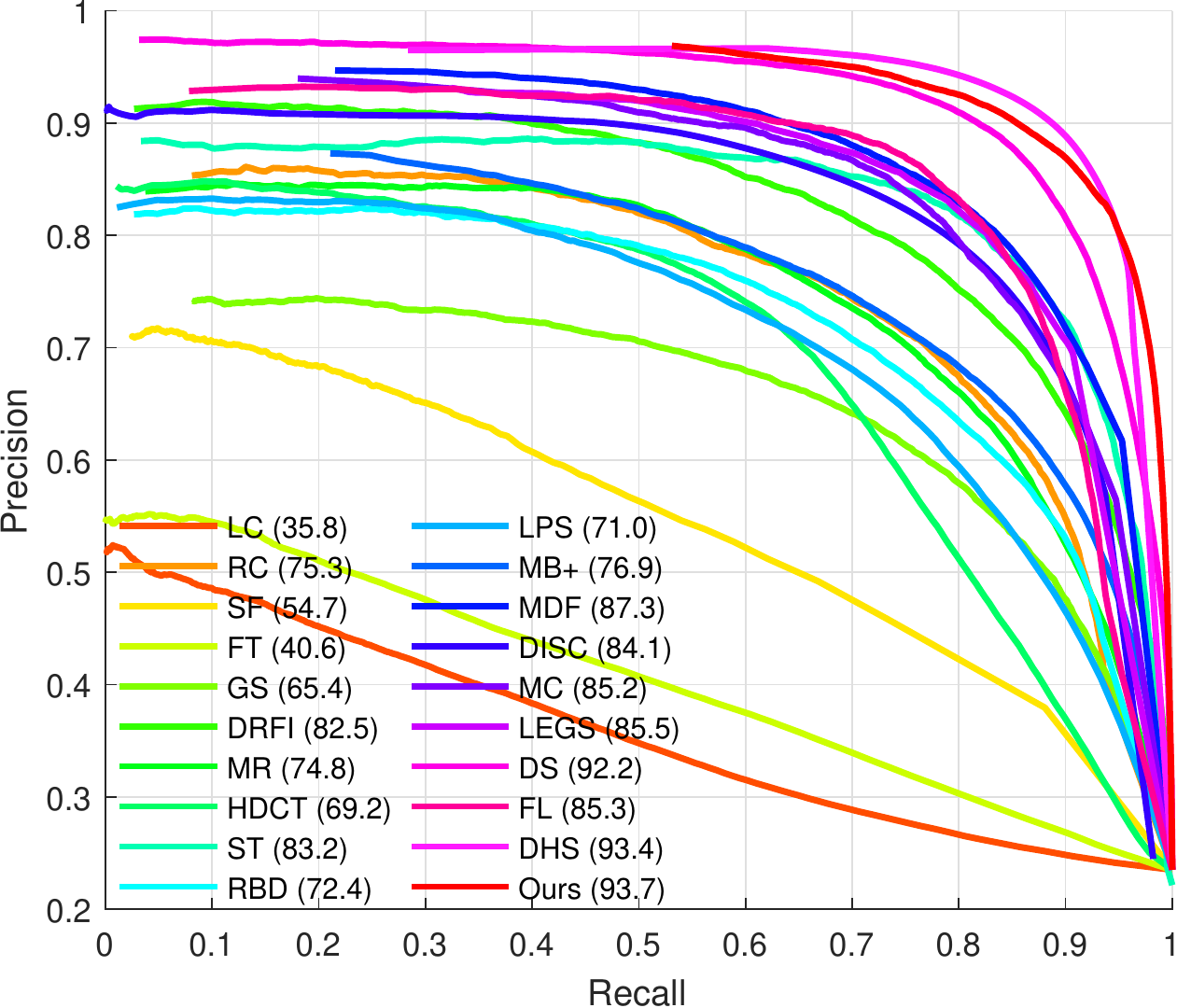}\hspace{3mm}
\includegraphics[width=0.33\linewidth,trim = 0mm 0mm 0mm 0mm, clip]{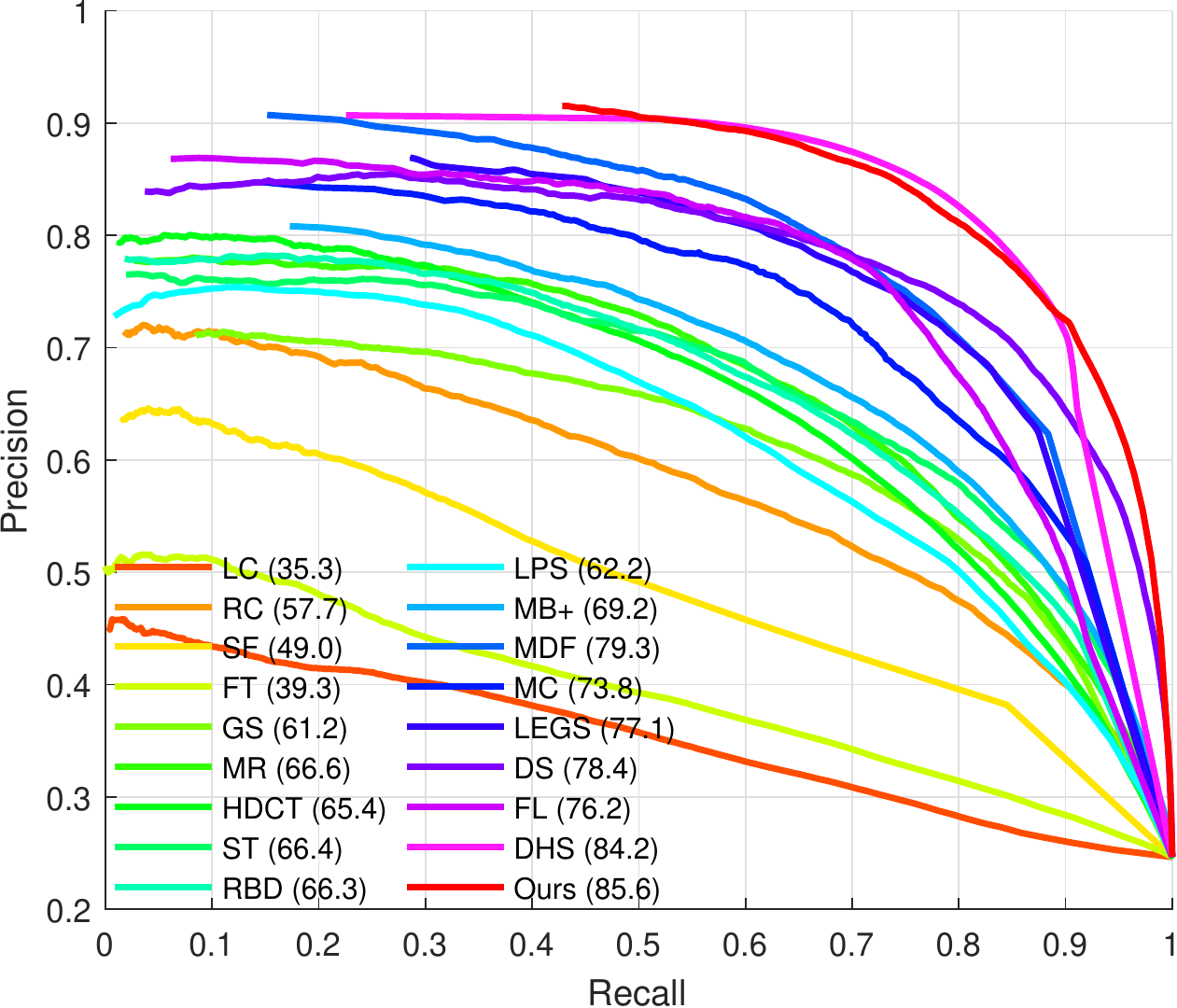}\hspace{2mm} \vspace{1 mm} \\
%\hspace{-3mm}\rotatebox{90}{\hspace{26mm}\textbf{DUT-ORMON}}  &\hspace{-0.8cm}
\includegraphics[width=0.33\linewidth,trim = 0mm 0mm 0mm 0mm, clip]{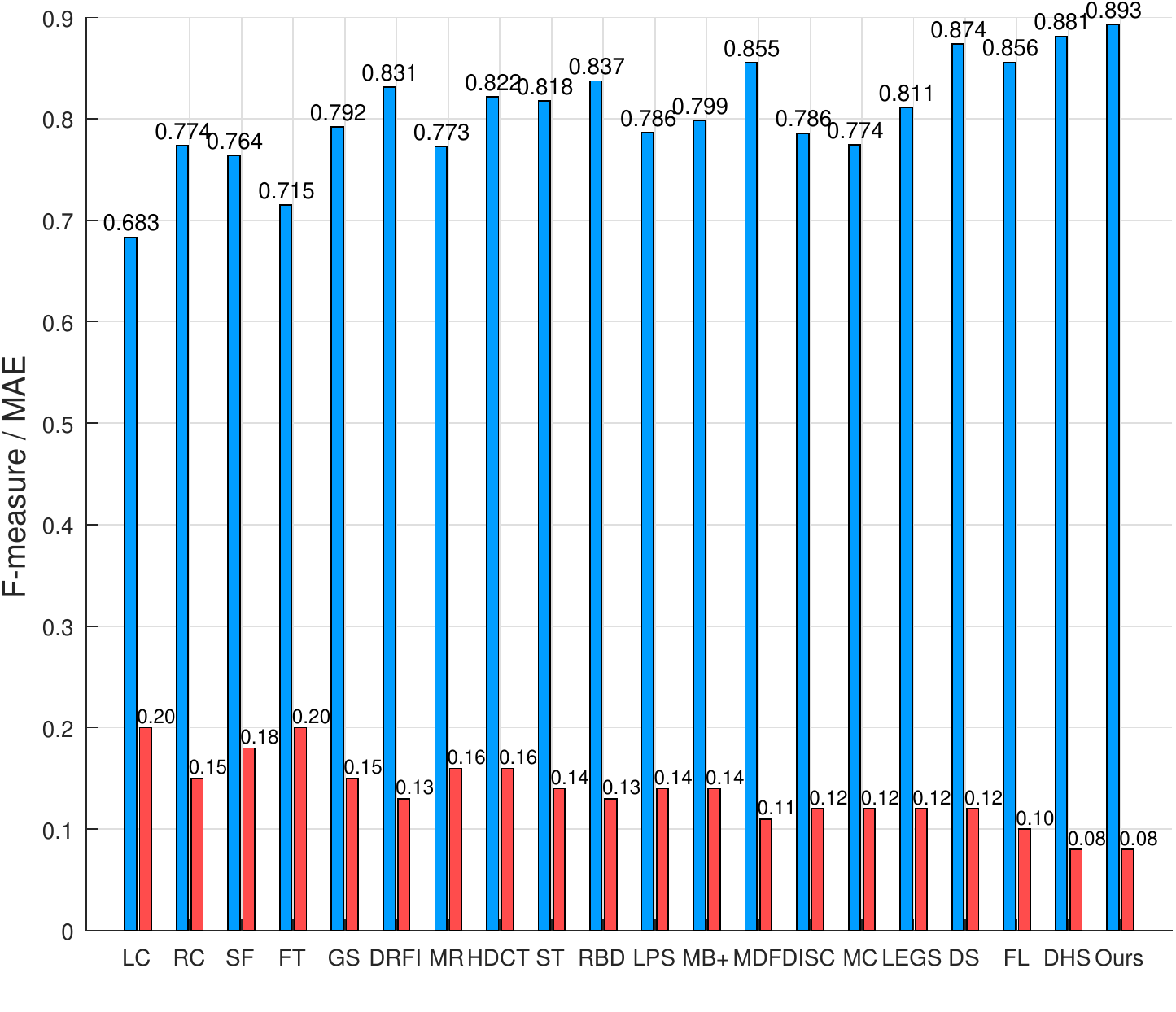}\hspace{3mm}
\includegraphics[width=0.33\linewidth,trim = 0mm 0mm 0mm 0mm, clip]{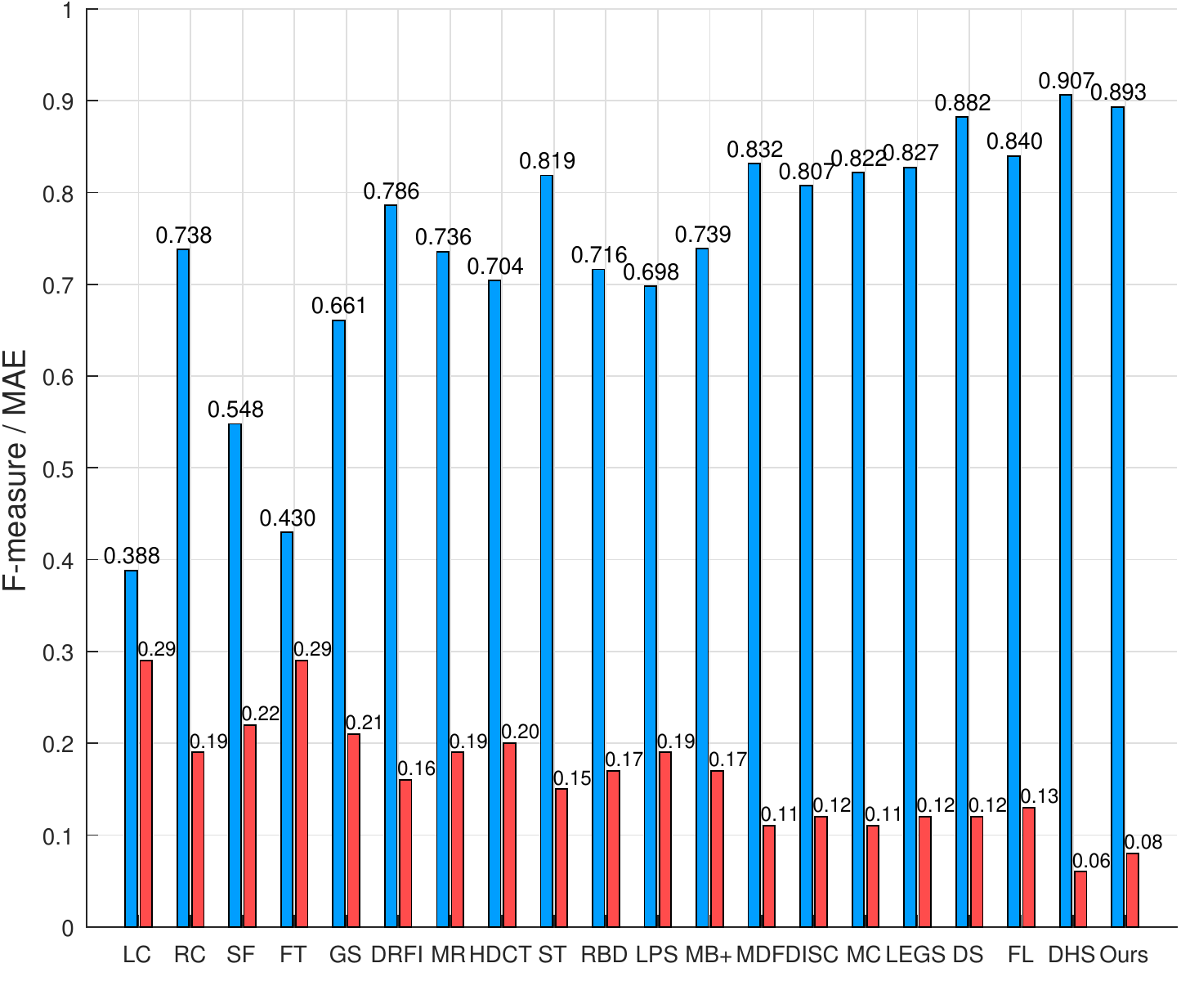}\hspace{3mm}
\includegraphics[width=0.33\linewidth,trim = 0mm 0mm 0mm 0mm, clip]{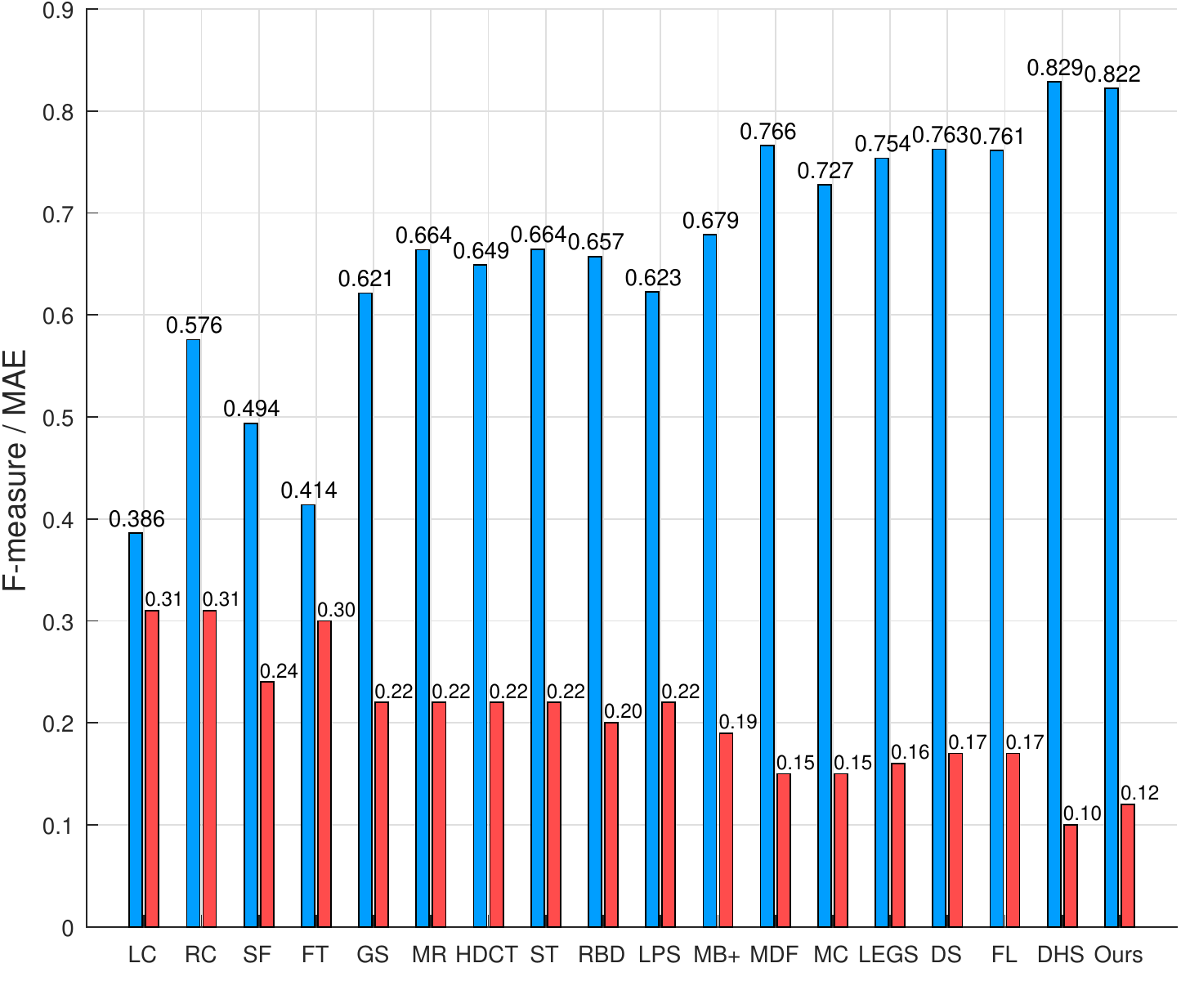}\hspace{2mm}\vspace{1 mm}\\
\begin{minipage}[b]{0.33\linewidth}
  \centering
  \centerline{\footnotesize{\textbf{SED2}}}\medskip
\end{minipage}
\begin{minipage}[b]{0.33\linewidth}
  \centering
  \centerline{\footnotesize{\textbf{ECSSD}}}\medskip
\end{minipage}
\begin{minipage}[b]{0.33\linewidth}
  \centering
  \centerline{\footnotesize{\textbf{Pascal-S}}}\medskip
\end{minipage}\\
\hfill
\vspace{-5 mm}
\end{tabular}
\caption{Comparison with state-of-the-art methods on six benchmark datasets. For each dataset, the first row shows the PR curves and the second row shows the F-measure and MAE. The numbers in the PR curves denote the AUC. Best viewed in color.}	
\label{fig:comparison}
\end{center}
\vspace{0mm}
\end{figure*}

\begin{figure*}[!ht] %\footnotesize
\begin{center}
\includegraphics[width=1\linewidth,trim = 0mm 0mm 0mm 0mm, clip]{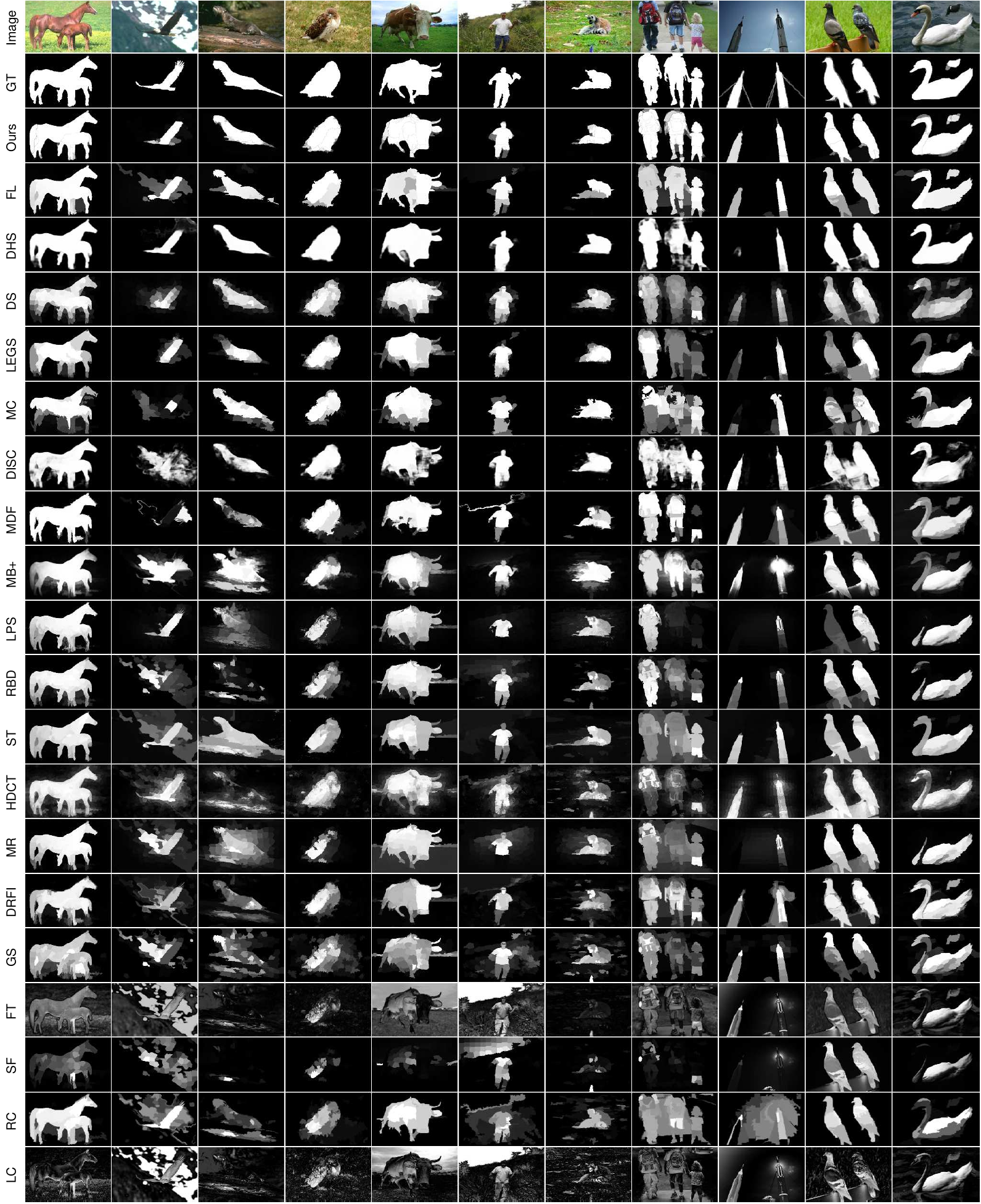}
\vspace{-2mm}
\caption{Qualitative comparison with state-of-the-art methods. We can see that our method locates salient objects more accurately and preserves object boundaries better. Background noise is strongly suppressed and the objects are highlighted uniformly.}
\label{fig:visual}
\end{center}
\vspace{0mm}
\end{figure*}

\subsection{Setup}

We randomly sample 4000 images from DUT-OMRON~\cite{yang2013saliency} dataset and 5000 images from MSRA10K~\cite{liu2007learning, achanta2009frequency, cheng2011global} dataset as training set and then evaluate our method on the following six benchmark datasets: ECSSD~\cite{yan2013hierarchical}, DUT-OMRON~\cite{yang2013saliency}, JuddDB~\cite{borji2015salient}, SED2~\cite{alpert2012image}, THUR15K~\cite{cheng2014salientshape} and Pascal-S~\cite{li2014secrets}. Note that the DUT-OMRON has 5168 images and we only evaluate on the remaining 1168 images that are not included in the training set. We also evaluate our method on two benchmark RGB-D saliency datasets: RGBD1000~\cite{peng2014rgbd} and NJU2000~\cite{ju2014depth}. All results are got from the benchmark of Borji \textit{et.al.}~\cite{borji2015benchmark} or generated using authors' code.

We evaluate the performance using precision-recall (PR) curves, F-measure and mean absolute error (MAE). The saliency maps are first normalized to $[0, 255]$, and then the precision and recall are computed by binarizing them with 256 thresholds and comparing them with ground truth. The PR curves are computed by averaging them on each dataset. The F-measure considers both precision and recall which is computed as:
\begin{equation}
F_\beta = \frac{(1+\beta^2)Precision \times Recall}{\beta^2 Precision + Recall},
\end{equation}
we set $\beta^2=0.3$ as most previous works~\cite{achanta2009frequency, cheng2011global} to emphasize the precision. The final F-measure is the maximal $F_\beta$ computed by 256 precision-recall pairs in the PR curves~\cite{borji2015benchmark}. The MAE directly measures the mean absolute difference between saliency map and ground truth,
\begin{equation}
MAE = \frac{1}{W\times H}\sum\limits_{x=1}^{W}\sum\limits_{y=1}^{H}|S(x,y)-GT(x,y)|
\end{equation}

\begin{figure}[t] %\footnotesize
\begin{center}
\begin{tabular}{ccc} % 42 40 28
\hspace{0mm}
\includegraphics[width=0.45\linewidth,trim = 0mm 0mm 0mm 0mm, clip]{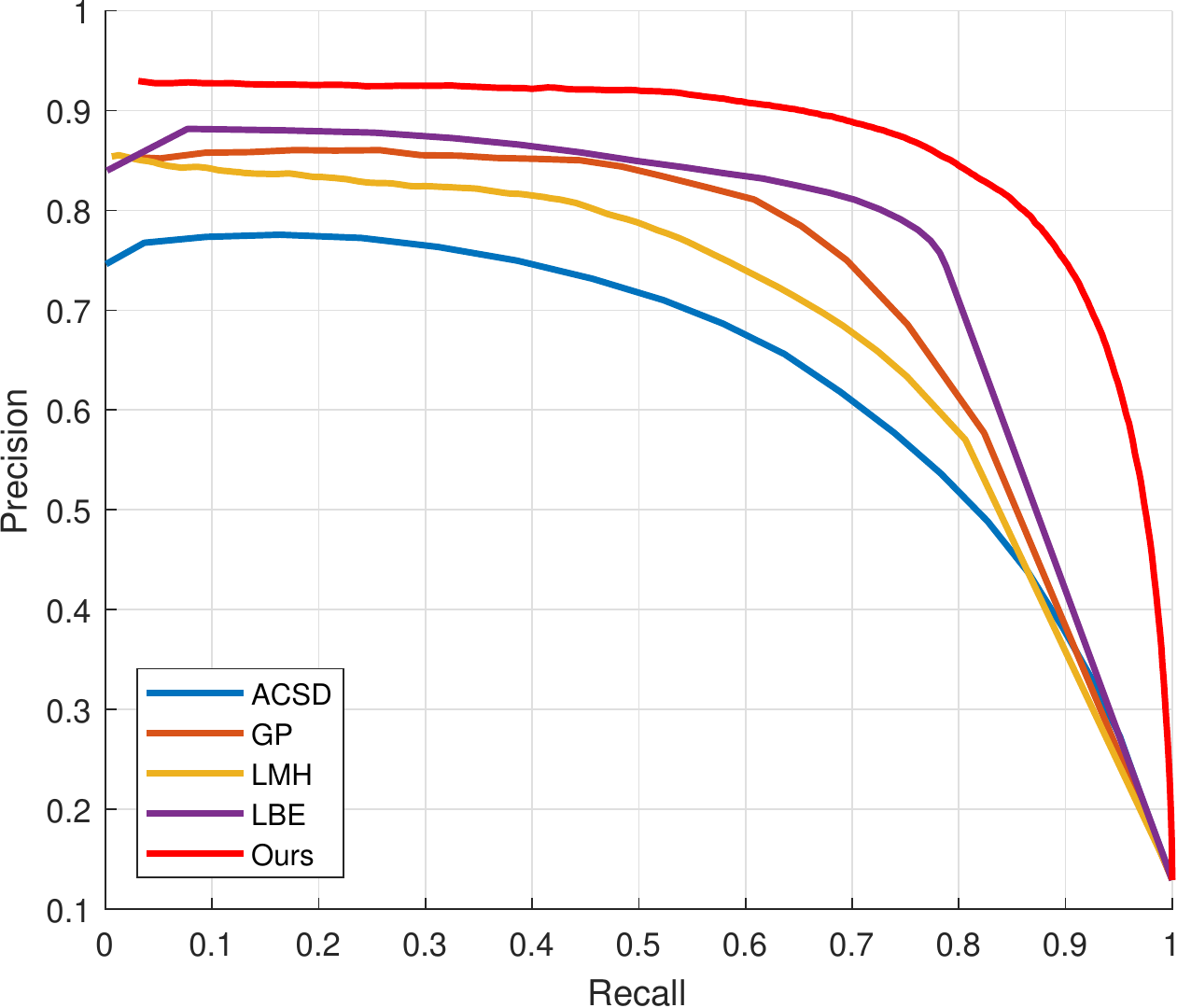}\hspace{3mm}
\includegraphics[width=0.45\linewidth,trim = 0mm 0mm 0mm 0mm, clip]{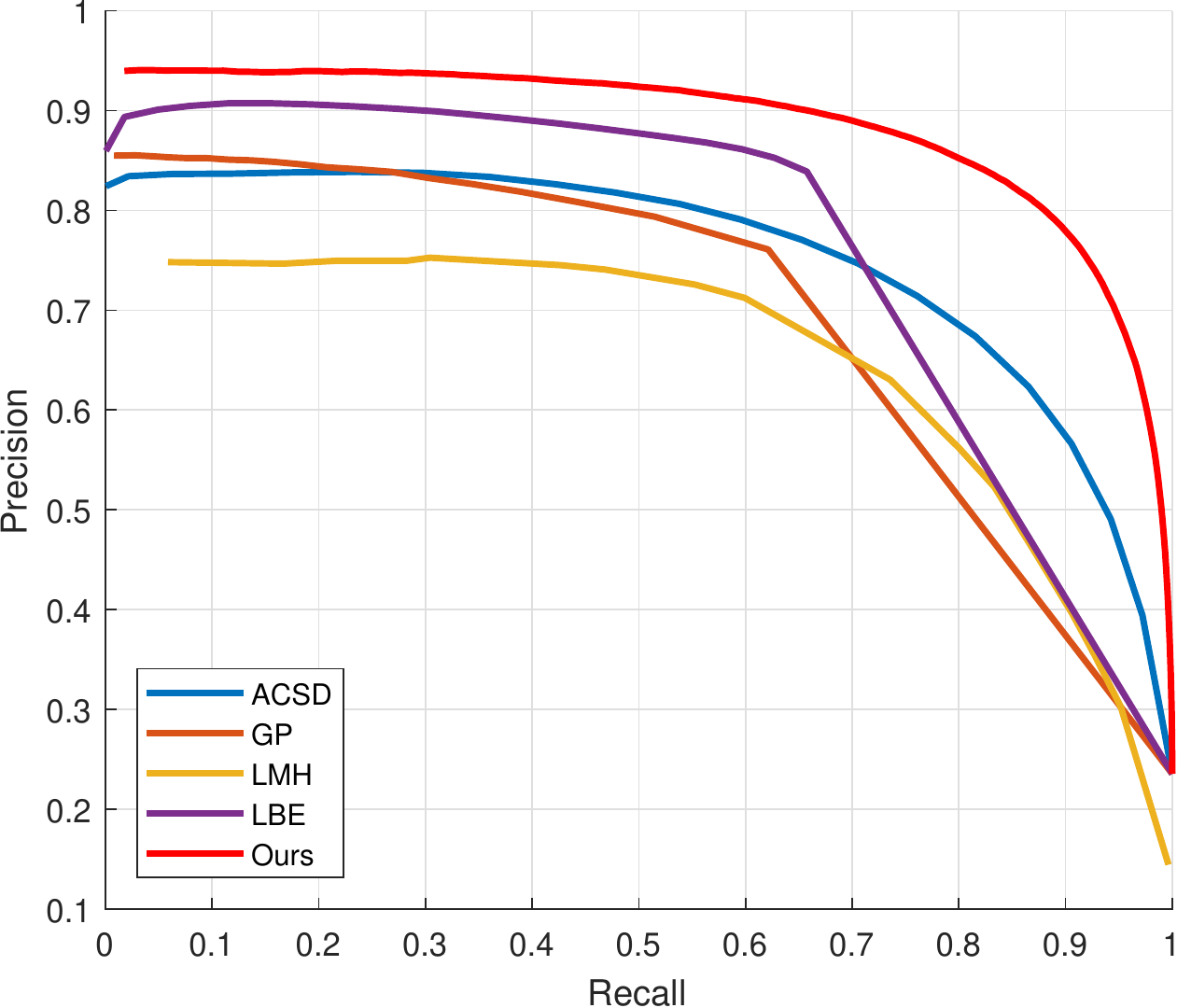}\hspace{3mm}
\vspace{-1 mm} \\
\begin{minipage}[b]{0.45\linewidth}
  \centering
  \centerline{\footnotesize{RGBD1000}}\medskip
\end{minipage}
\begin{minipage}[b]{0.45\linewidth}
  \centering
  \centerline{\footnotesize{NJU2000}}\medskip
\end{minipage}\\
\hfill
\vspace{-5 mm}
\end{tabular}
\caption{Comparison with state-of-the-art methods on two benchmark RGB-D saliency datasets. Best viewed in color.}
\label{fig:RGBD}
\end{center}
\vspace{-2mm}
\end{figure}

\begin{table}[!t]
  \centering
  \caption{Training data of state-of-the-art methods.}
% Table generated by Excel2LaTeX from sheet 'Sheet1'
\begin{tabular}{c|l}

\Xhline{1.0pt}
    Method & Training Data \\
\Xhline{1.0pt}
       MDF~\cite{li2015visual} & 2,500 images from MSRA-5000 \\
    \hline
      DISC~\cite{chen2015disc} & 9,000 images from MSRA10K \\
    \hline
        MC~\cite{zhao2015saliency} & 8,000 images from MSRA10K \\
    \hline
\multicolumn{ 1}{c|}{} &  3,000 images from the MSRA-5000 dataset and           \\

\multicolumn{ 1}{c|}{LEGS~\cite{wang2015deep}} & 340 images from the Pascal-S dataset.      \\
\multicolumn{ 1}{c|}{} & Both horizontal reflection and rescaling (±5\%) are applied  \\
    \hline
        DS~\cite{li2016deepsaliency} & leave-one-out strategy, using other 7 datasets for training \\
    \hline
        DHSNet~\cite{liu2016dhsnet} & 6,000 from MSRA10K and 3,500 from DUT-OMRON \\
    \hline
      OURS & 4,000 from DUT-OMRON and 5,000 from MSRA10K \\
\Xhline{1.0pt}

\end{tabular}  
  \label{tab:trainingdata}%
\end{table}%

\subsection{Comparison with State-of-the-art Methods}
We compare our method with state-of-the-art methods, including traditional methods: LC~\cite{zhai2006visual}, RC~\cite{cheng2011global}, SF~\cite{perazzi2012saliency}, FT~\cite{achanta2009frequency}, GS~\cite{wei2012geodesic}, DRFI~\cite{jiang2013drfi} MR~\cite{yang2013saliency}, HDCT~\cite{kim2014salient}, ST~\cite{liu2014saliency}, RBD~\cite{zhu2014saliency}, LPS~\cite{li2015inner}, MB+~\cite{zhang2015minimum}, and CNN based methods: MDF~\cite{li2015visual}, DISC~\cite{chen2015disc}, MC~\cite{zhao2015saliency}, LEGS~\cite{wang2015deep}, DS~\cite{li2016deepsaliency}, DHSNet~\cite{liu2016dhsnet} and our preliminary conference method FL~\cite{fl2016}. For CNN-based methods, we also list the training data they used in Table~\ref{tab:trainingdata}. MDF~\cite{li2015visual} uses less training data, DS~\cite{li2016deepsaliency} uses much more training data, and for other methods, we use comparable training data. \fref{fig:comparison} shows PR-curves, F-measure and MAE on six benchmark datasets. 
We can see that our method outperforms other methods and our preliminary conference method by a large margin. 
For the state-of-the art multi-scale method DHSNet~\cite{liu2016dhsnet}, we achieve comparable performance. For PR curves, our method outperforms DHSNet on all datasets by $2.6\%$ on average. For F-measure, our method outperforms DHSNet on JuddDB, THUR15K and SED2 datasets, but fails on ECSSD and Pascal-S dataset. For MAE, we are inferior to DHSNet by $0.026$ on average.

Note that DS~\cite{li2016deepsaliency} is a multi-task framework which detects salient object and object boundaries simultaneously, our method outperforms DS~\cite{li2016deepsaliency} at all 6 datasets, especially on datasets with complex scenes, such as DUT-OMRON, JuddDB and Pascal-S, which shows that our method better takes advantages of edges.
Note that our network is trained on parts of DUT-OMRON and MSRA10K dataset, we apply the trained network to other 5 datasets without fine-tuning, the results still outperform others by a large margin, which shows that our method has strong generalization ability. \fref{fig:visual} shows the qualitative comparison with state-of-the-art methods, we can see that our method preserves edges well and suppresses most background noise.

\begin{figure}[t] %\footnotesize
\begin{center}
\includegraphics[width=0.98\linewidth,trim = 0mm 0mm 0mm 0mm, clip]{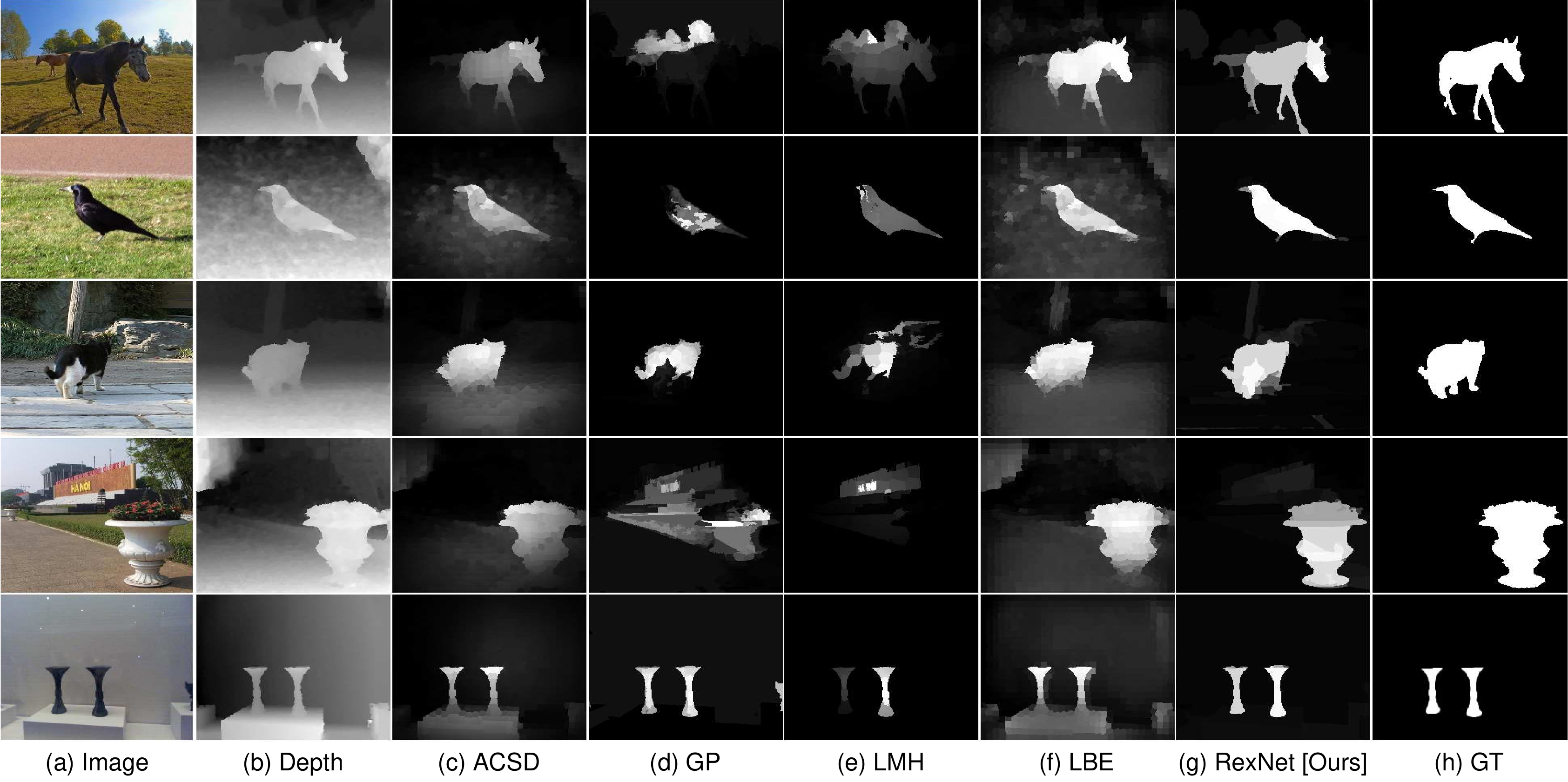}
\vspace{-3mm}
\caption{Qualitative comparison with state-of-the-art methods on RGB-D datasets. Our method can not only locate salient object accurately, but also preserve edges, thus highlighting the whole object uniformly and suppressing background noise.}
\label{fig:depth_compare}
\end{center}
\vspace{0mm}
\end{figure}

\subsection{Evaluation on RGB-D Saliency Datasets}
We compare our method with state-of-the-art RGB-D saliency methods: ACSD~\cite{ju2014depth}, GP~\cite{ren2015exploiting}, LMH~\cite{peng2014rgbd} and LBE~\cite{Feng2016Local}. \fref{fig:RGBD} shows the comparison of PR-curves. Our method significantly outperforms other methods, especially in the region of high recall. The main reason of our performance is that our method can not only locate salient object accurately, but also preserve edges, thus saliency map of our method are with high precision and high recall. \fref{fig:depth_compare} also shows the qualitative comparison with state-of-the-art RGB-D methods.

\begin{table*}[htbp]
  \centering
  \caption{Evaluation of all components on six benchmark datasets with F-measure and AUC. The final result $S$ always performs better than all components, which shows that all the components are complementary and our method is effective.}
    \begin{tabular}{c|cc|cc|cc|cc|cc|cc}
\Xhline{1.0pt}
          & \multicolumn{2}{c|}{JuddDB} & \multicolumn{2}{c|}{DUT-OMRON} & \multicolumn{2}{c|}{THUR15K} & \multicolumn{2}{c|}{SED2} & \multicolumn{2}{c|}{ECSSD} & \multicolumn{2}{c}{Pascal-S} \\
          & $F_\beta$ & AUC   & $F_\beta$ & AUC   & $F_\beta$ & AUC   & $F_\beta$ & AUC   & $F_\beta$ & AUC   & $F_\beta$ & AUC \\
\Xhline{1.0pt}
    $S_S$ &      0.490 &      0.457 &      0.722 &      0.720 &      0.706 &      0.710 &      0.849 &      0.851 &      0.851 &      0.901 &      0.768 &      0.806 \\

    $S_E$ &      0.515 &      0.464 & 0.771 &      0.728 & 0.734 &      0.696 & 0.882 &      0.861 &      0.858 &      0.864 &      0.789 &      0.802 \\

    $S_C$ & 0.534 & 0.508 &      0.762 & 0.770 &      0.721 & 0.717 &      0.877 & 0.883 & 0.874 & 0.914 & 0.799 & 0.836 \\

      $S$ & {\bf 0.556} & {\bf 0.545} & {\bf 0.789} & {\bf 0.803} & {\bf 0.761} & {\bf 0.779} & {\bf 0.893} & {\bf 0.918} & {\bf 0.893} & {\bf 0.937} & {\bf 0.822} & {\bf 0.856} \\

\Xhline{1.0pt}
    \end{tabular}%
  \label{tab:eva1}%
\end{table*}%

\begin{table*}[htbp]
  \centering
  \caption{Result of each branch and their fusion in \textit{ContextNet}.}
    \begin{tabular}{c|cc|cc|cc|cc|cc|cc}
\Xhline{1.0pt}
          & \multicolumn{2}{c|}{JuddDB} & \multicolumn{2}{c|}{DUT-OMRON} & \multicolumn{2}{c|}{THUR15K} & \multicolumn{2}{c|}{SED2} & \multicolumn{2}{c|}{ECSSD} & \multicolumn{2}{c}{Pascal-S} \\
          & $F_\beta$ & AUC   & $F_\beta$ & AUC   & $F_\beta$ & AUC   & $F_\beta$ & AUC   & $F_\beta$ & AUC   & $F_\beta$ & AUC \\
\Xhline{1.0pt}
Branch 1 &      0.402 &      0.366 &      0.529 &      0.510 &      0.533 &      0.510 &      0.749 &      0.780 &      0.639 &      0.643 &      0.599 &      0.596 \\

  Branch 2 &      0.416 &      0.381 &      0.525 &      0.507 &      0.557 &      0.540 &      0.691 &      0.728 &      0.692 &      0.719 &      0.622 &      0.622 \\

  Branch 3 &      0.447 &      0.423 &      0.564 &      0.563 &      0.600 &      0.601 &      0.705 &      0.737 &      0.751 &      0.801 &      0.678 &      0.713 \\

  Branch 4 &      0.490 &      0.457 &      0.692 &      0.710 &      0.686 &      0.695 &      0.802 &      0.854 &      0.836 &      0.891 &      0.756 &      0.798 \\

    Fusion & {\bf 0.534} & {\bf 0.508} & {\bf 0.762} & {\bf 0.770} & {\bf 0.721} & {\bf 0.717} & {\bf 0.877} & {\bf 0.883} & {\bf 0.874} & {\bf 0.914} & {\bf 0.799} & {\bf 0.836} \\
	
\Xhline{1.0pt}
    \end{tabular}%
  \label{tab:branches}%
\end{table*}%

\subsection{Ablation Studies}
In this subsection, we conduct experiments to verify the effectiveness of  each component of our method.

\textbf{Network Components.} We first evaluate the components of the proposed network by outputting the intermediate results of our network and analyzing their performance.
Table~\ref{tab:eva1} shows the comparison of all components: $S_S$, $S_E$, $S_C$ and the final saliency map $S$ on six benchmark datasets. To better demonstrate the comparison with numerical results, we use Area Under Curve (AUC) which measures the area under the PR-curve to represent PR-curve criterion. 
We can see that the final result $S$ outperforms all components, which shows that all the components are complementary and our method is effective.

\textbf{Branches of \textit{ContextNet}.} We evaluate the effectiveness of branches of \textit{ContextNet}. Table~\ref{tab:branches} shows the results of each branch and the fusion results on six benchmark datasets. We can see that, commonly, the branches of deeper layers achieve better performance, and the final fusion result is the best, which demonstrates that our method makes full use of features at each branch.

\textbf{Edge Loss.} We evaluate the effectiveness of Edge Loss by comparing with networks without Edge Loss. Table~\ref{tab:eva_Edge} shows the results of \textit{ContextNet} on six benchmark datasets. With the Edge Loss, the performance is better since the Edge Loss can preserve edges better and so the saliency map of \textit{ContextNet} are more uniform.

\begin{table*}[htbp]
  \centering
  \caption{Results of \textit{ContextNet} with and without Edge Loss. With the Edge Loss, the performance is better.}
    \begin{tabular}{c|cc|cc|cc|cc|cc|cc}
\Xhline{1.0pt}
          & \multicolumn{2}{c|}{JuddDB} & \multicolumn{2}{c|}{DUT-OMRON} & \multicolumn{2}{c|}{THUR15K} & \multicolumn{2}{c|}{SED2} & \multicolumn{2}{c|}{ECSSD} & \multicolumn{2}{c}{Pascal-S} \\
          & $F_\beta$ & AUC   & $F_\beta$ & AUC   & $F_\beta$ & AUC   & $F_\beta$ & AUC   & $F_\beta$ & AUC   & $F_\beta$ & AUC \\
\Xhline{1.0pt}
w/o Edge Loss &      0.524 &      0.494 &      0.750 &      0.744 &      0.715 &      0.703 &      0.873 &      0.865 &      0.865 &      0.903 &      0.789 &      0.822 \\

w/ Edge Loss & {\bf 0.534} & {\bf 0.508} & {\bf 0.762} & {\bf 0.770} & {\bf 0.721} & {\bf 0.717} & {\bf 0.877} & {\bf 0.883} & {\bf 0.874} & {\bf 0.914} & {\bf 0.799} & {\bf 0.836} \\

\Xhline{1.0pt}
    \end{tabular}%
  \label{tab:eva_Edge}%
\end{table*}%

\textbf{Comparison with fusing features.} The proposed \textit{ContextNet} fuses saliency maps of each branch to get the final result. To evaluate the effectiveness, we compare with method which fuses features to predict saliency map. We concatenate features of each branch to predict saliency map. Table~\ref{tab:fusingfeatures} shows the result of \textit{ContextNet} with fusing features and fusing maps. We can see that our method outperforms method which fuses features. This is benefited from the deep supervision in each branch which makes full use of features at different levels.

\begin{table*}[htbp]
  \centering
  \caption{Comparison with fusing features. Our proposed fusing maps method outperforms method which fuses features.}
    \begin{tabular}{c|cc|cc|cc|cc|cc|cc}
\Xhline{1.0pt}
          & \multicolumn{2}{c|}{JuddDB} & \multicolumn{2}{c|}{DUT-OMRON} & \multicolumn{2}{c|}{THUR15K} & \multicolumn{2}{c|}{SED2} & \multicolumn{2}{c|}{ECSSD} & \multicolumn{2}{c}{Pascal-S} \\
          & $F_\beta$ & AUC   & $F_\beta$ & AUC   & $F_\beta$ & AUC   & $F_\beta$ & AUC   & $F_\beta$ & AUC   & $F_\beta$ & AUC \\
\Xhline{1.0pt}
Fusing Features &      0.520 &      0.486 &      0.734 &      0.724 &      0.704 &      0.686 &      0.873 &      0.871 &      0.855 &      0.887 &      0.776 &      0.805 \\

Fusing Maps & {\bf 0.534} & {\bf 0.508} & {\bf 0.762} & {\bf 0.770} & {\bf 0.721} & {\bf 0.717} & {\bf 0.877} & {\bf 0.883} & {\bf 0.874} & {\bf 0.914} & {\bf 0.799} & {\bf 0.836} \\

\Xhline{1.0pt}
    \end{tabular}%
  \label{tab:fusingfeatures}%
\end{table*}%

\begin{figure}[!t] %\footnotesize
\begin{center}
%\begin{tabular}{cccc}
\begin{tabular}{ccc} % 42 40 28
%\hspace{-3mm}\rotatebox{90}{\hspace{38mm}\textbf{ECSSD}}  &\hspace{-0.8cm}\hspace{2mm}
\hspace{-2mm}
\includegraphics[width=0.45\linewidth,trim = 0mm 0mm 0mm 0mm, clip]{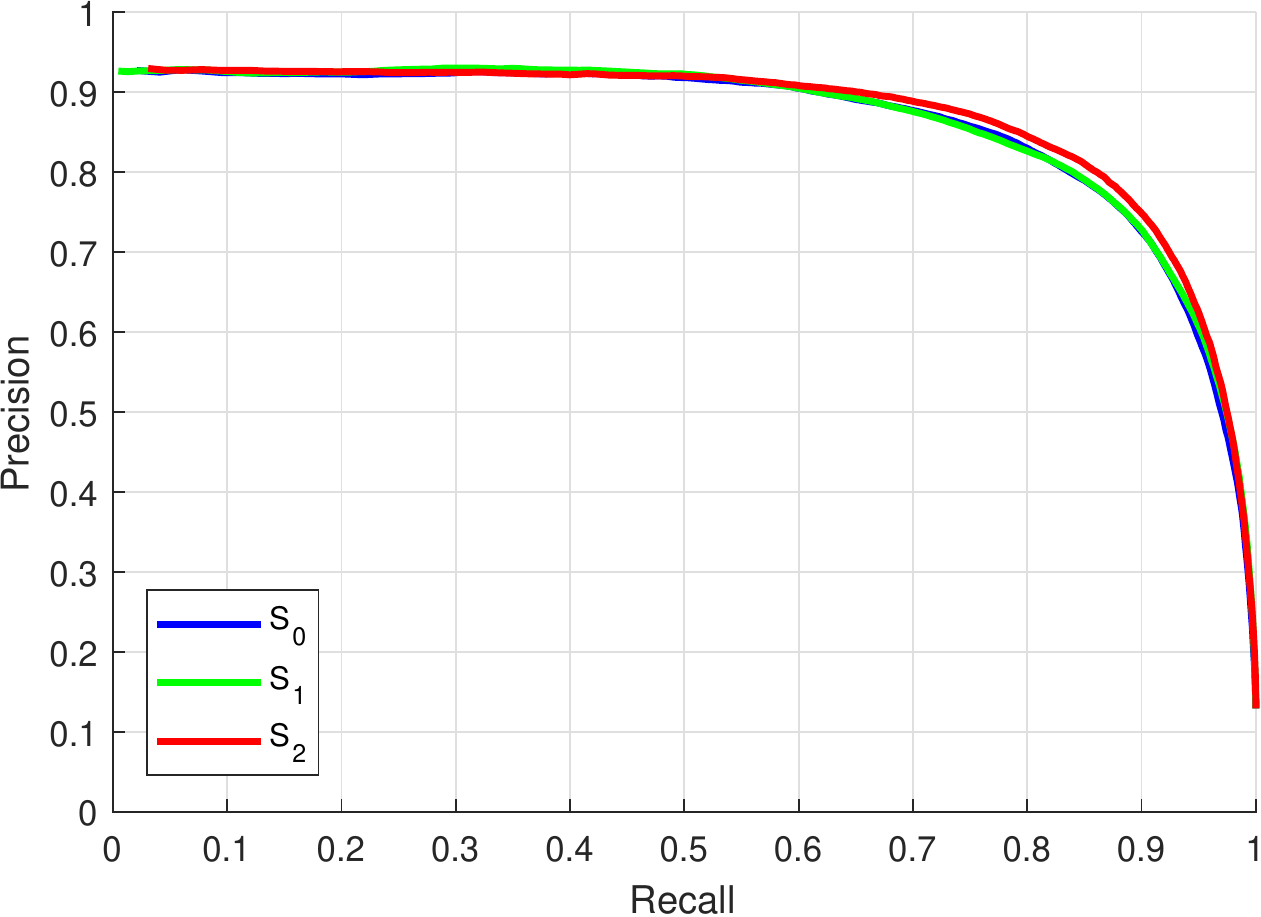}\hspace{3mm}
\includegraphics[width=0.45\linewidth,trim = 0mm 0mm 0mm 0mm, clip]{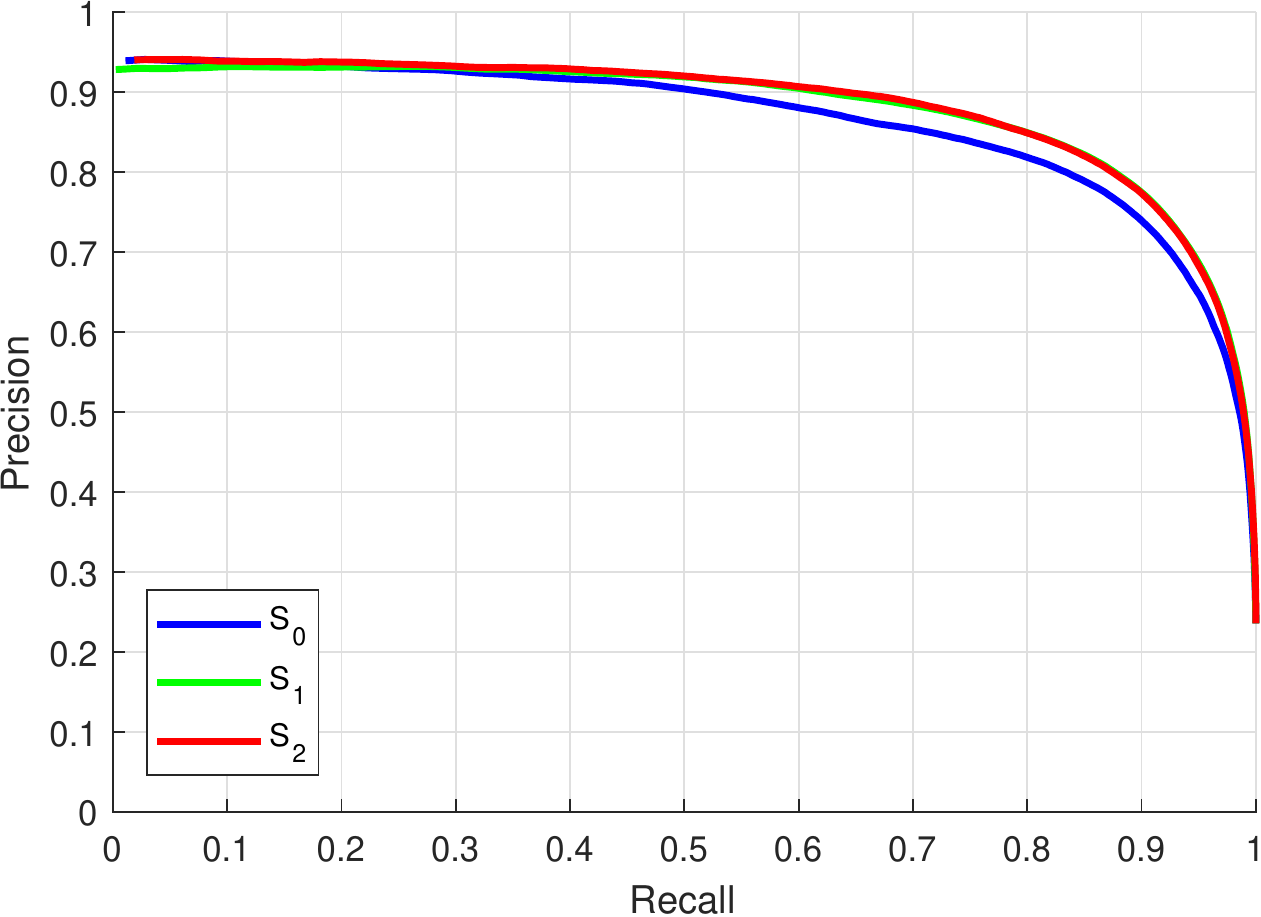}\hspace{3mm}
 \vspace{-1 mm} \\

\begin{minipage}[b]{0.45\linewidth}
  \centering
  \centerline{\footnotesize{RGBD1000}}\medskip
\end{minipage}
\begin{minipage}[b]{0.45\linewidth}
  \centering
  \centerline{\footnotesize{NJU2000}}\medskip
\end{minipage}\\
\hfill
\vspace{-5 mm}
\end{tabular}

\caption{Evaluate the effectiveness of depth refinement. Our depth refinement improves the performance mainly at the region with high recall, which is essential important for the final performance. Best viewed in color.}
\label{fig:depth}
\end{center}
\vspace{-2mm}
\end{figure}

\textbf{Depth Refinement.} For the RGB-D saliency datasets, we evaluate the effectiveness of depth refinement. We show the comparison of PR-curves with and without depth refinement in \fref{fig:depth}. Experimental results show that the depth refinement improve the performance significantly especially in the region with high precision and high recall.

\textbf{Speed.} We compare the speed with other region-based CNN methods. Our method is much faster since we deal with regions under end-to-end Fast R-CNN framework, while other region-based CNN methods forward network for each region. Table~\ref{tab:speed} shows the comparison of performance and running time, the experiment is conduct on ECSSD dataset~\cite{yan2013hierarchical}, it contains 1000 test images, we test on this dataset with a single NVIDIA GeForce GTX TITAN GPU and report the average time per image. We compare with MC~\cite{zhao2015saliency} and LEGS~\cite{wang2015deep} using the authors' public code. Our method takes $0.75s$ for each image, including $0.4s$ for segmenting image into regions using superpixel and edges and only $0.35s$ for network forwarding. Our method takes less time while achieving better performance.

% Table generated by Excel2LaTeX from sheet 'Sheet3'
\begin{table}[!t]
  \centering
  \caption{Performance and speed comparison with other region-based CNN methods. Our method takes $0.4s$ for segmenting image into regions, and only $0.35s$ for network forwarding. Our method takes less time while achieving better performance.}
    \begin{tabular}{c|c|c|c}
\Xhline{1.0pt}
          & $F_\beta$   & AUC & Time (s) \\
\Xhline{1.0pt}
    \textit{RexNet} [Ours]   & 0.893 & 0.937& 0.40 + 0.35\\
    MC~\cite{zhao2015saliency}     & 0.822 & 0.852& 1.63 \\
    LEGS~\cite{wang2015deep}   & 0.827 & 0.855 & 2.27 \\
\Xhline{1.0pt}
    \end{tabular}%
  \label{tab:speed}%
\end{table}%

\subsection{Failure Cases}
\begin{figure}[!t] %\footnotesize
\begin{center}
\includegraphics[width=0.98\linewidth,trim = 0mm 0mm 0mm 0mm, clip]{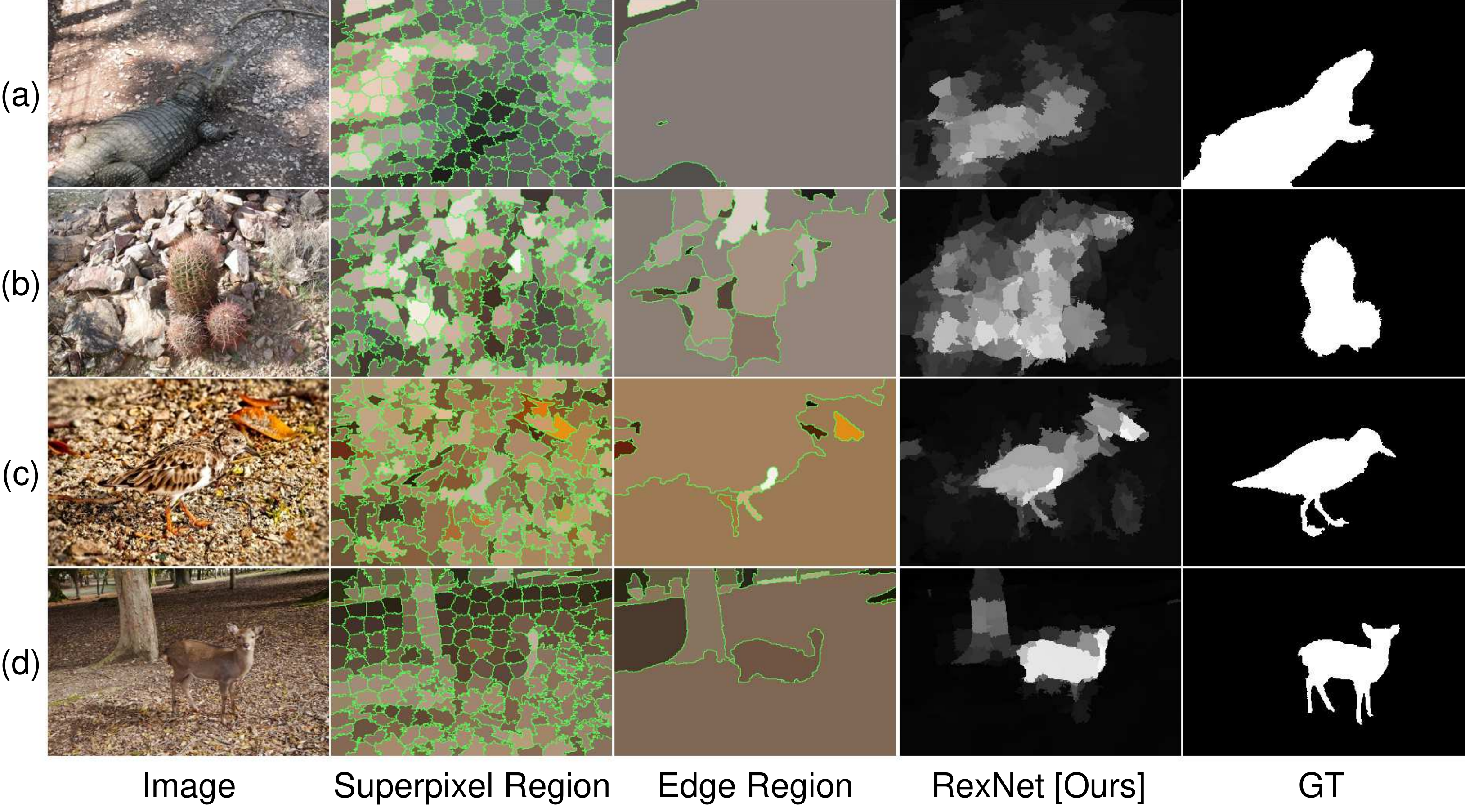}
\vspace{-3mm}
\caption{Some failure cases of our method. These images are with extreme low-contrast scenes, which makes it difficult to segment into correct regions, thus influencing the final results. (a, b) both superpixel and edge segmentation fail, the result is bad. (c, d) the boundary between object and background is a bit clearer, thus the result is much better than (a) and (b).}
\label{fig:failure_case}
\end{center}
\vspace{-8mm}
\end{figure}

Our proposed framework achieves state-of-the-art performance. However, as the \textit{RegionNet} is based on the segmentation of images, when the image is with extreme low contrast and the boundary between object and background is blurry, the segmentation may fail and thus influencing the final performance. \fref{fig:failure_case} shows some failure examples. These images are all in scene with low contrast, when both superpixel and edge segmentation fail, the performance decreases much.
Note that in \fref{fig:failure_case} (c) and (d), though the scene is low-contrast, the boundary between object and background is a bit clearer, thus the result is much better than \fref{fig:failure_case} (a) and (b).

\section{Conclusion}\label{sec:conclusion}
In this paper, we propose \textit{RexNet} %to address limits of existing CNN based methods. 
%motivated by the drawbacks of existing CNN based methods, we propose an edge-preserving and multi-scale contextual neural network 
which generates saliency map end-to-end and with sharp object boundaries. In the proposed framework, image is first segmented into two scales of complementary regions: superpixel regions and edge regions. The network then generates saliency score of regions end-to-end and context in multiple layers are considered to fuse with region saliency scores. The proposed \textit{RexNet} achieves both clear detection boundary and multi-scale contextual robustness simultaneously for the first time, thus achieves an optimized performance. We also extend the proposed framework to RGB-D saliency detection by depth refinement. Experiments on benchmark RGB and RGB-D datasets demonstrate that the proposed method achieves state-of-the-art performance.

\section*{Acknowledgement}
This work was supported by National Natural Science Foundation of China (No. 61171113) and National Key Basic Research Program of China (No. 2016YFB0100900).

% Can use something like this to put references on a page
% by themselves when using endfloat and the captionsoff option.
\ifCLASSOPTIONcaptionsoff
  \newpage
\fi

% trigger a \newpage just before the given reference
% number - used to balance the columns on the last page
% adjust value as needed - may need to be readjusted if
% the document is modified later
%\IEEEtriggeratref{8}
% The "triggered" command can be changed if desired:
%\IEEEtriggercmd{\enlargethispage{-5in}}

% references section

% can use a bibliography generated by BibTeX as a .bbl file
% BibTeX documentation can be easily obtained at:
% http://mirror.ctan.org/biblio/bibtex/contrib/doc/
% The IEEEtran BibTeX style support page is at:
% http://www.michaelshell.org/tex/ieeetran/bibtex/
%\bibliographystyle{IEEEtran}
% argument is your BibTeX string definitions and bibliography database(s)
%\bibliography{IEEEabrv,../bib/paper}
%
% <OR> manually copy in the resultant .bbl file
% set second argument of \begin to the number of references
% (used to reserve space for the reference number labels box)

\bibliographystyle{IEEEtran}
\bibliography{mybibfile}

% insert where needed to balance the two columns on the last page with
% biographies
%\newpage

% You can push biographies down or up by placing
% a \vfill before or after them. The appropriate
% use of \vfill depends on what kind of text is
% on the last page and whether or not the columns
% are being equalized.

%\vfill

% Can be used to pull up biographies so that the bottom of the last one
% is flush with the other column.
%\enlargethispage{-5in}

% that's all folks
\end{document}